\documentclass[sigconf]{acmart}
\usepackage{caption}
\usepackage{multicol}
\usepackage{multirow}
\usepackage{subcaption}
\AtBeginDocument{%
  \providecommand\BibTeX{{%
    \normalfont B\kern-0.5em{\scshape i\kern-0.25em b}\kern-0.8em\TeX}}}

\copyrightyear{2020}
\acmYear{2020}
\setcopyright{acmcopyright}\acmConference[MM '20]{Proceedings of the 28th ACM International Conference on Multimedia}{October 12--16, 2020}{Seattle, WA, USA}
\acmBooktitle{Proceedings of the 28th ACM International Conference on Multimedia (MM '20), October 12--16, 2020, Seattle, WA, USA}
\acmPrice{15.00}
\acmDOI{10.1145/3394171.3413934}
\acmISBN{978-1-4503-7988-5/20/10}

\begin{document}

\title{MOR-UAV: A Benchmark Dataset and Baselines for Moving Object Recognition in UAV Videos}

\author{Murari Mandal, Lav Kush Kumar, Santosh Kumar Vipparthi}
\email{murarimandal.cv@gmail.com, lavkushkumarmnit@gmail.com, skvipparthi@mnit.ac.in}
\affiliation{
 Vision Intelligence Lab,
   \institution{Malaviya National Institute of Technology Jaipur, INDIA}}


\begin{abstract}
Visual data collected from Unmanned Aerial Vehicles (UAVs) has opened a new frontier of computer vision that requires automated analysis of aerial images/videos. However, the existing UAV datasets primarily focus on object detection. An object detector does not differentiate between the moving and non-moving objects. Given a real-time UAV video stream, how can we both localize and classify the moving objects, i.e. perform moving object recognition (MOR)? The MOR is one of the essential tasks to support various UAV vision-based applications including aerial surveillance, search and rescue, event recognition, urban and rural scene understanding.To the best of our knowledge, no labeled dataset is available for MOR evaluation in UAV videos. Therefore, in this paper, we introduce MOR-UAV, a large-scale video dataset for MOR in aerial videos. We achieve this by labeling axis-aligned bounding boxes for moving objects which requires less computational resources than producing pixel-level estimates. We annotate 89,783 moving object instances collected from 30 UAV videos, consisting of 10,948 frames in various scenarios such as weather conditions, occlusion, changing flying altitude and multiple camera views. We assigned the labels for two categories of vehicles (car and heavy vehicle). Furthermore, we propose a deep unified framework MOR-UAVNet for MOR in UAV videos. Since, this is a first attempt for MOR in UAV videos, we present 16 baseline results based on the proposed framework over the MOR-UAV dataset through quantitative and qualitative experiments. We also analyze the motion-salient regions in the network through multiple layer visualizations. The MOR-UAVNet works online at inference as it requires only few past frames. Moreover, it doesn’t require predefined target initialization from user. Experiments also demonstrate that the MOR-UAV dataset is quite challenging.
\end{abstract}



\maketitle

\section{Introduction}
Visual data captured from unmanned aerial vehicles (UAVs) is used in numerous applications for military, industry and consumer market space. It has opened a new frontier of computer vision, i.e., aerial vision, which requires analysis and interpretation of images and videos gathered from UAVs either in real-time or in post-event study. Some of the applications include aerial surveillance, search and rescue, exploration, urban planning, industrial and agricultural monitoring, action/event recognition, sports analysis and scene understanding \cite{yu2019unmanned,xia2018dota}.\par

\begin{figure*}[t]
\centering
     \includegraphics[width=0.9\textwidth ]{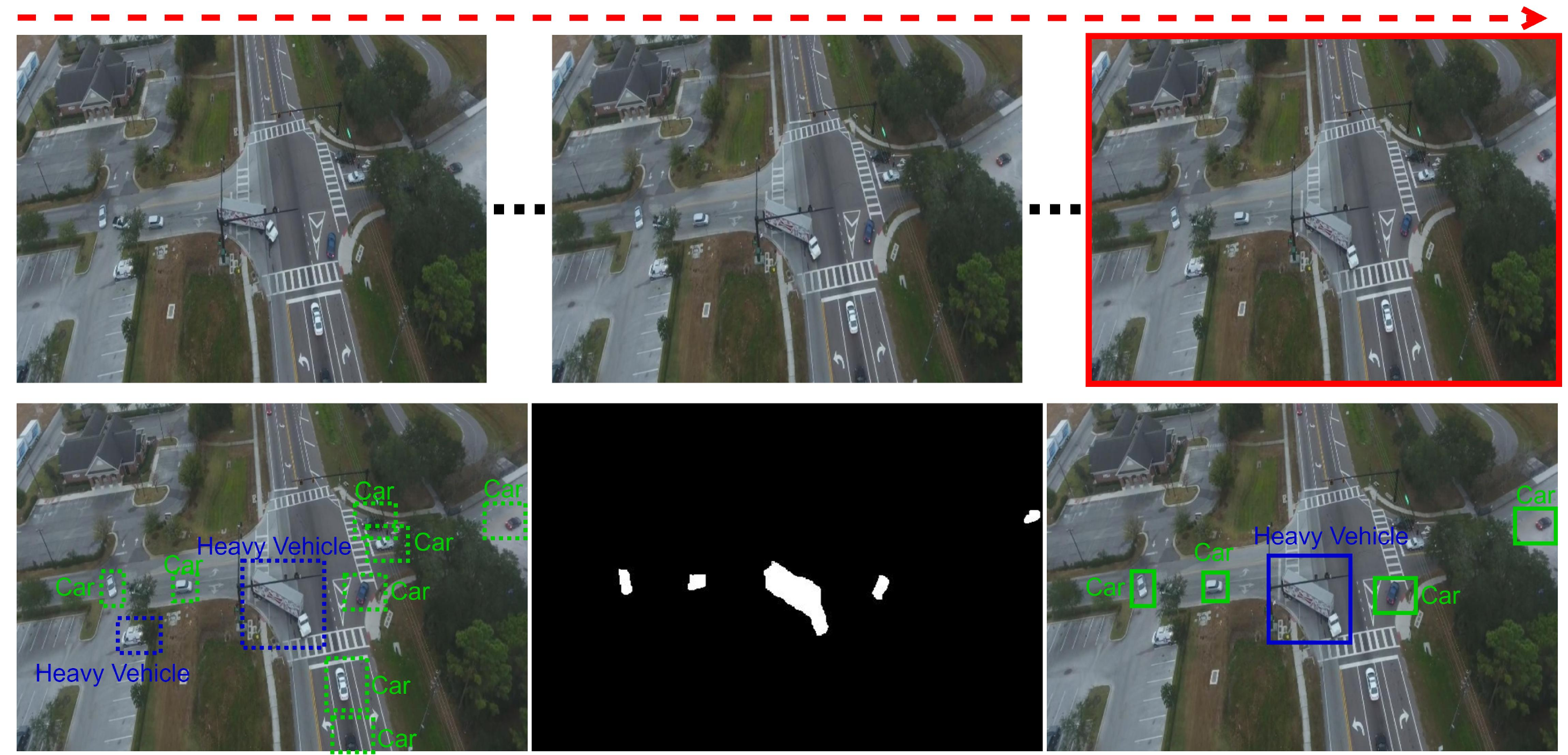}
	\hfill\small \hspace{6cm} (a) OD \hspace{4cm} (b) MOD \hspace{4cm} (c) MOR
	\caption{Difference between the three tasks: object detection (OD), moving object detection (MOD) and MOR is depicted. The OD algorithm detects all (moving and non-moving) object instances. Whereas, in MOD, only the pixel-wise changes are identified in a class-agnostic manner. The proposed MOR method simultaneously detects and classifies the moving objects in the video. The same is shown in this figure.}
	\label{fig:figure1}
  \label{fig:figure1}
\end{figure*}

In addition to the existing datasets and algorithms~\cite{deng2009imagenet,he2015spatial,he2016deep,sandler2018mobilenetv2,girshick2014rich,kong2017ron,shrivastava2016training,liu2016ssd,everingham2010pascal,sermanet2013overfeat,lin2014microsoft,chen2018encoder,he2017mask,xu2018youtube,maninis2018video,barekatain2017okutama,singh2018sniper,hosang2017learning,law2018cornernet,pont20172017,bao2018cnn,cheng2017segflow,jain2017fusionseg,koh2017primary,papazoglou2013fast,song2018pyramid,ochs2013segmentation,wang2014cdnet,mehta2020hidegan,marotirao2019challenges,dwivedi20203d,Mandal_2020_WACV} for analysis of images/videos captured in regular view, researchers in the literature have also developed numerous UAV datasets for some of the fundamental tasks such as detection and tracking. The Campus~\cite{robicquet2016learning} and CARPK~\cite{hsieh2017drone} videos are captured in specific locations such as campus or parking lot. The UAV123\cite{mueller2016benchmark} was developed for low-altitude object tracking in both real and simulated scenes. Similarly, the Okutama~\cite{barekatain2017okutama} provides annotations for human action recognition in aerial views. More recently, several datasets were presented for object detection~\cite{yu2019unmanned,xia2018dota,barekatain2017okutama,liu2015fast,razakarivony2016vehicle,zhu2019visdrone,xu2019dac}
and tracking~\cite{yu2019unmanned, hsieh2017drone,zhu2018vision,du2019visdrone} in unconstrained aerial scenarios. These datasets have accelerated deep learning research in UAV vision-based applications. However, no labeled dataset is available in the literature for one of the low-level tasks of moving object recognition (MOR), i.e. simultaneous localization and classification of moving objects in a video frame. The task of MOR is different from both object detection and visual tracking. In both these tasks, the methods do not differentiate between the moving and non-moving objects. Furthermore, MOR is even different from moving object detection (MOD) which performs pixel-wise binary segmentation in each frame. The difference between object detection (OD), MOD and MOR are demonstrated in Figure~\ref{fig:figure1}. MOR has widespread applications in intelligent visual surveillance, intrusion detection, agricultural monitoring, industrial site monitoring, detection-based tracking, autonomous vehicles, etc. More specifically, the MOR algorithm can recognize the movements of different classes of objects such as people, vehicles, etc. in real-world scenarios and this information can further be used in high-level decision making such as anomaly detection and selective target tracking. Thus, there is a need for a comprehensive UAV benchmark video dataset for MOR in unconstrained scenarios.\par

\begin{figure*}[t]
\centering
	\includegraphics[width=0.9\textwidth ]{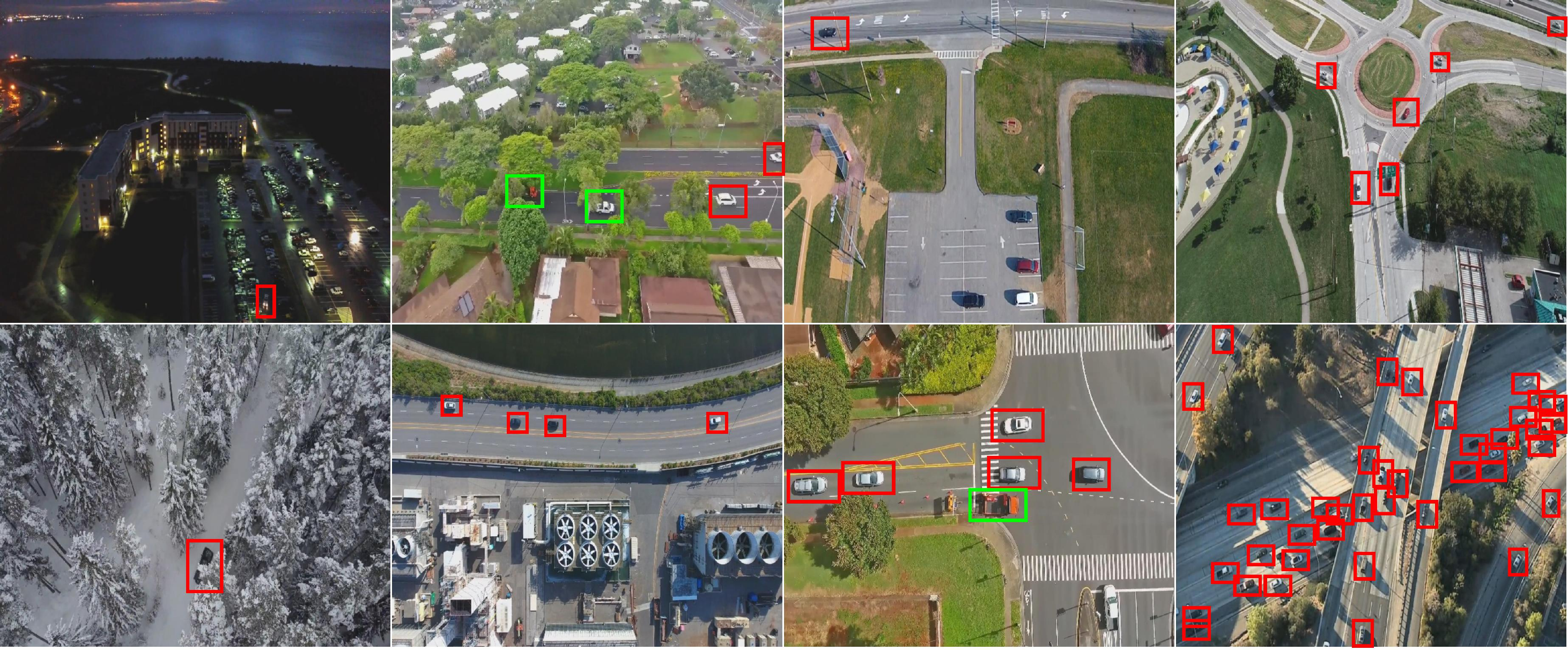}
	\caption{Sample video frames taken from MOR-UAV. The videos are collected from different scenarios including night time, occlusion, camera motion, weather conditions, camera views, etc. The red and green boxes denote two object classes car and heavy-vehicle respectively.}
	\label{figure2}
\end{figure*}

To advance the MOR research in aerial videos, this paper introduces a largescale challenging UAV moving object recognition benchmark dataset named MOR-UAV. The MOR-UAV consists of $89,783$ moving object instances annotated in 30 videos comprising of 10,948 frames. The videos are captured in various challenging scenarios such as night time, occlusion, camera motion, weather conditions, camera views, etc. Moreover, no constraints are enforced to maintain the same camera sensor and platform for all videos. This puts the onus on the MOR algorithm to robustly recognize the moving objects in the wild. The altitude and camera views also vary in different video sequences. All the moving objects are manually annotated with bounding boxes along with the object category. In this paper, the objects of interest are two types of vehicles: \textit{cars} and \textit{heavy vehicles}. A sample set of video frames from MOR-UAV dataset is depicted in Figure~\ref{figure2}. The complete dataset with annotations will be made publicly available in the future.\par

We also present an online deep unified framework named MOR-UAVNet for MOR evaluation in the newly constructed dataset. The proposed framework retains the property of online inference by using only the recent history frames in live streaming videos. This paper makes the following contributions: 
\begin{itemize}
  \item We introduce a fully annotated dataset, MOR-UAV for the fundamental task of MOR in unconstrained UAV videos. To the best the author's knowledge, this is the first aerial/UAV video dataset for simultaneous localization and classification of moving objects.
  \item We propose a novel deep learning framework MOR-UAVNet to simultaneously localize and classify the moving objects in the UAV videos. The visualizations of different convolutional layers in MOR-UAVNet is qualitatively analyzed for a better understanding of the network functionality for MOR.
  \item Based on the new MOR-UAVNet framework, we evaluate 16 baseline models for MOR evaluation over MOR-UAV. We also depict the qualitative results for MOR in UAV videos.
\end{itemize}

\section{MOR-UAV Dataset}
\subsection{Comparison with Existing UAV Datasets}
Advancement in deep learning algorithms and the availability of largescale labeled datasets has fueled the progress in important applications in several domains. Some of the low-level tasks in computer vision include image classification~\cite{deng2009imagenet,he2015spatial,he2016deep,sandler2018mobilenetv2}, object detection~\cite{lin2017feature,linfocalpami,liu2016ssd,7485869,lin2019real,zhang2018mfr}, semantic segmentation~\cite{chen2018encoder,he2017mask}, video object segmentation~\cite{maninis2018video,xu2018youtube,pont20172017}, motion detection~\cite{ochs2013segmentation,wang2014cdnet,mandal2018candid,yang2019unsupervised,mandal20193dfr,akilan2019sendec,mandal2018antic} and visual tracking~\cite{an2019new,valmadre2017end,fan2019siamese}. Although many challenging applications are presented to the researchers in UAV based computer vision community. However, limited labeled datasets~\cite{yu2019unmanned,xia2018dota,xu2019dac,zhu2019visdrone,razakarivony2016vehicle,liu2015fast,barekatain2017okutama,du2019visdrone,mueller2016benchmark,zhu2018vision} are available for accelerating the improvement and evaluation of various vision tasks. Recently, numerous datasets have been constructed with annotations for object detection and visual tracking. Muellerl~\cite{mueller2016benchmark} presented a tracking dataset recorded from drone cameras to evaluate the ability of single object trackers to tackle camera movements, illumination variations, and object scale changes. Moreover, several video clips were recorded for pedestrian behavior analysis from an aerial view in~\cite{robicquet2016learning}. Hsieh et al. ~\cite{hsieh2017drone} augmented a dataset for vehicle counting in parking lots. Similarly, for object detection in aerial views, several datasets have been presented over time. Some of the most comprehensive and challenging datasets are the DOTA~\cite{xia2018dota}, UAVDet~\cite{yu2019unmanned}, VisDrone~\cite{zhu2018vision,zhu2019visdrone} and Dac-sdc~\cite{xu2019dac}. It has led to rapid advancement in the development of specialized object detectors~\cite{ding2019learning,liang2019small,wu2019orsim,mandal2019sssdet,liu2018detection,mandal2019avdnet,bazi2018convolutional,gong2019context,wu2019delving} for aerial images. Similarly, VisDrone~\cite{du2019visdrone}, UAVDet~\cite{yu2019unmanned} and UAV123~\cite{mueller2016benchmark} have provided annotated data for bench-marking visual tracking in UAV videos. Several tracking algorithms~\cite{wang2017robust,pan2019multi} have been evaluated over these datasets to advance research in aerial object tracking. Furthermore, an extension of DOTA was recently made available in~\cite{waqas2019isaid} to also facilitate instance segmentation in aerial images as well.\par

Few researchers~\cite{berker2017feature,lezki2018joint} have utilized the videos from VIVID~\cite{collins2005open} and UAV123 to self-annotate a few frames for moving object detection (MOD). Others~\cite{li2016multi} have collected UAV videos for specialized purposes. However, to the best of our knowledge, no labeled dataset is available in the literature for MOR. The proposed MOR-UAV is a first large-scale dataset with bounding-box labels which can be used as the benchmark for both MOR and MOD in UAV videos. The dataset consists of videos from numerous unconstrained scenarios, resulting in better generalization to unseen videos. The detailed comparison of the proposed MOR-UAV with other datasets in the literature is summarized in Table~\ref{table1}. 

\begin{table}[t]
\centering
\caption{Comparison of MOR-UAV with other largescale UAV datasets. Det: Detection, T: Visual tracking, Act: Action recognition, MOR: Moving object recognition}
\begin{tabular}{c c c} 
 \hline 
 Dataset & Tasks & Labeled Moving Objects\\ 
 \hline 
 VisDrone~\cite{zhu2018vision}& Det, T & No\\ 
 DOTA~\cite{xia2018dota} & Det & No  \\ 
 UAV123~\cite{mueller2016benchmark}&  T & No\\ 
 UAVDT~\cite{yu2019unmanned} & Det, T & No \\
 Okutama~\cite{barekatain2017okutama} & Det, Act & No \\   
 Dac-sdc~\cite{xu2019dac} &  Det & No \\ 
  \textbf{MOR-UAV} &  \textbf{MOR} &  \textbf{Yes} \\
 \hline
\end{tabular}
\label{table1}
\end{table}

\subsection{Data Collection and Annotation}
The MOR-UAV dataset comprises of $30$ videos collected from multiple video recordings captured with a UAV. Locations in highways, flyovers, traffic intersections, urban areas and agricultural regions are collected for analysis. These videos represent various scenarios including occlusion, nighttime, weather changes, camera motion, changing altitudes, different camera views, and angles. The videos are recorded at $30$ frames per second (fps) and the resolution varies from $1280\times 720$ to $1920\times 1080$. The average, min, max lengths of the sequences are $364.93, 64$ and $1,146$ respectively.\par

\begin{figure}[t]
\centering
	\includegraphics[width=0.5\textwidth ]{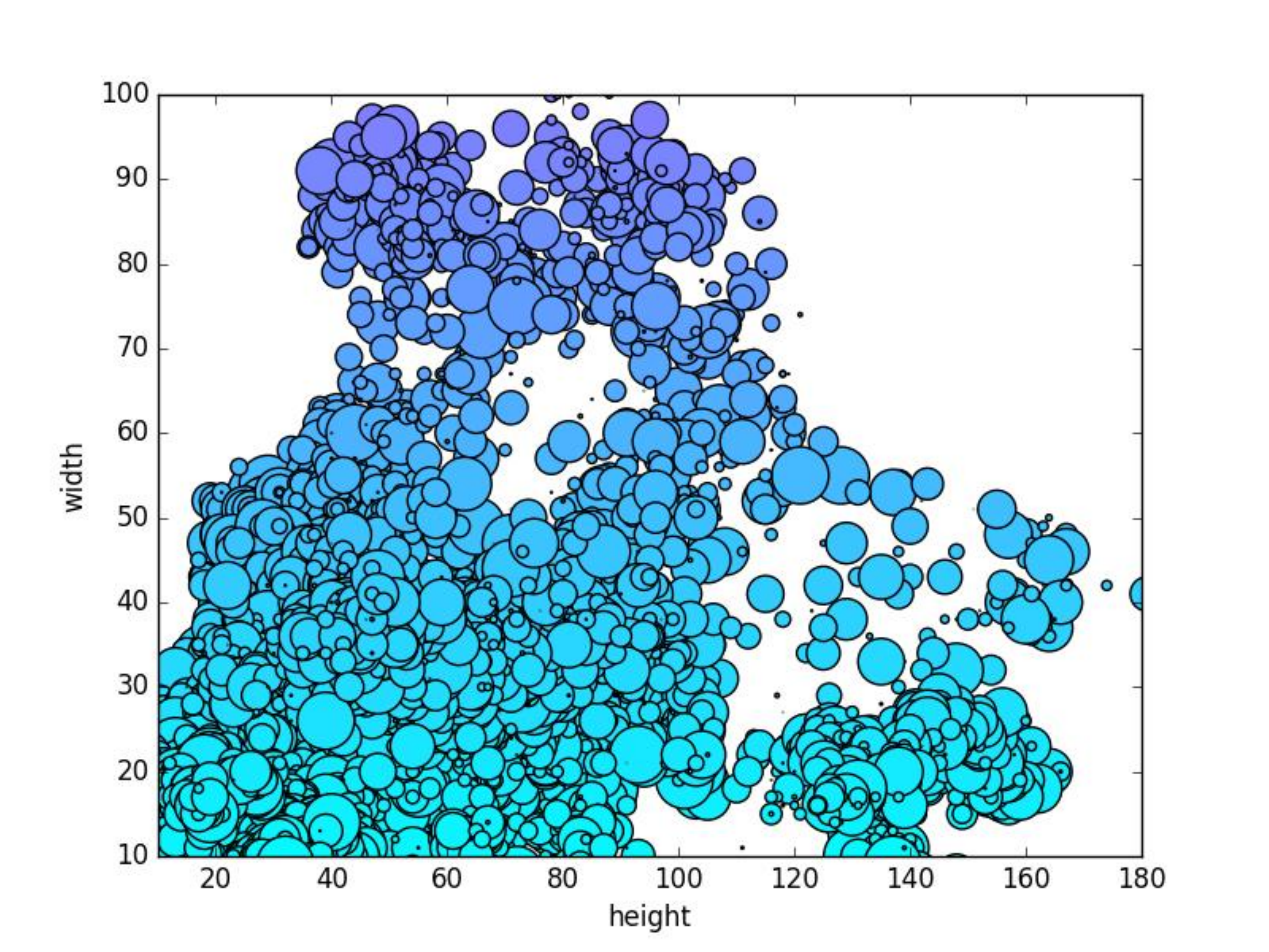}
	\hfill \small 
	\textit{
	(a) Avg. BB height = 29.01, avg. BB width = 17.64, min. BB height = 6, min BB width = 6, max. BB height = 181, max. BB width = 106\\
	(b) Avg. video sequence length = 364.93, min. video sequence length = 64, max. video sequence length = 1,146}
	\caption{The bounding-box (BB) height-width scatter-plot of all the object instances in MOR-UAV along with the complete dataset description. All the videos are normalized to the shape of $608\times 608\times 3$ for uniform analysis of the complete dataset.}
	\label{fig:figure3}
\end{figure}

The moving objects are labeled using the Yolo-mark\footnote{https://github.com/AlexeyAB/Yolo\_mark} tool. The bounding boxes are described with
$(x1, y1, x2, y2, c)$, where $(x1, y1)$ and $(x2, y2)$ are the top-left and bottom-right locations of the bounding box respectively. The object class is represented with $c$. About $10,948$ frames in the MOR-UAV dataset are annotated with approximately $89,783$ bounding boxes representing moving vehicles. These vehicles are categorized into two classes:\textit{car ($80,340$ bounding boxes)} and \textit{heavy vehicles ($9,443$ bounding boxes)}. Figure \ref{figure2} shows some sample frames with annotations in the dataset. The $average, min, max$ lengths of the bounding box heights are $29.011, 6, 181$, respectively. Similarly, $average, min, max$ lengths of the bounding box widths are $17.641, 6, 106$, respectively. The complete dataset details are depicted in Figure~\ref{fig:figure3}. 

\subsection{Dataset Attributes}
Some of the challenging attributes of the MOR-UAV dataset are as follows: 

\textbf{Variable object density.} The UAV videos are captured in both dense and sparsely populated regions. For example, a large number of vehicles are present in flyovers, parking lot and traffic signal intersections. Whereas, very few objects are present in forest, agricultural and other remote areas with complex backgrounds. Such diverse scenarios in terms of object density make it challenging for the MOR algorithms to obtain robust performance.\par

\textbf{Small and large object shapes.}
Due to high altitude of the UAVs, the objects appear very small. Thus, it is very difficult to accurately detect the motion features. Moreover, in some low altitude UAV videos, the objects appear reasonably large due to closer view. Thus, for the same class, both large, medium and small shapes of objects are present in the dataset. These multiscale object appearances challenge the researchers to design more generalizable algorithms.\par

\textbf{Sporadic camera motion.} In addition to the variable speed of object movements, the UAV camera speed and rotation is also unconstrained in the dataset. Sometimes the camera itself moves faster or rotates drastically. This makes it very difficult to differentiate between moving and non-moving objects. Moreover, sometimes it creates confusion between camera motion and actual object motion.\par

\textbf{Changes in the aerial view.} Changes in the camera orientations in UAV sometimes make the object appear from different side view angles or corner view in the video. Thus, multiple views of the objects appear in different videos or sometimes even in the same video.\par

Since the videos are collected from different open-source platforms, the exact details about the UAV altitude, camera angle, UAV speed, etc. are not available. Such unconstrained data collection makes the algorithm design more challenging and ensures robust performance in real-world UAV videos. Thus, the algorithms designed for MOR in this UAV video dataset should consider these practical factors. Moreover, the weather changes and nighttime videos present further challenges for accurate moving object recognition. 

\section{MOR-UAVNet Framework and Baseline Models}
\begin{figure*}[]
\centering
	\includegraphics[width=1\textwidth ]{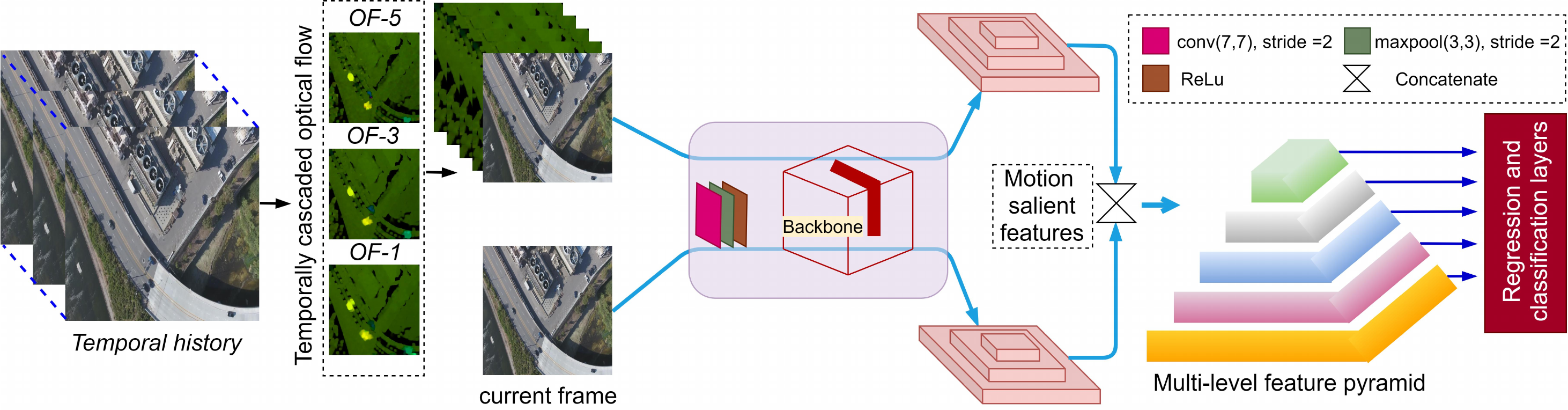}
	\caption{Schematic illustration of the proposed MOR-UAVNet framework for MOR in UAV videos. The motion saliency is estimated through cascaded optical flow computation at multiple stages in the temporal history frames. In this figure, optical flow between the current frame and the last ($OF-1$), third last ($OF-3$), fifth last frame ($OF-5$) is computed respectively. We then assimilate the salient motion features with the current frame. These assimilated features are forwarded through the ResNet backbone to extract spatial and temporal dimension aware features. Moreover, the base features from the current frame are also extracted to reinforce the semantic context of the object instances. These two feature maps are concatenated at matching scales to produce a feature map for motion encoding. Afterward, multi-level feature pyramids are generated. The dense bounding box and category scores are generated at each level of the pyramid. We use 5 pyramid levels in our experiments. \textit{This figure shows the MOR-UAVNetv1 model architecture}}
	\label{figure4}
\end{figure*}

We propose a deep unified framework MOR-UAVNet for simultaneous localization and classification of moving objects. The overall architecture of MOR-UAVNet is shown in Figure~\ref{figure4}. We discuss the functionality of the proposed network along with visualizations in the following subsections.

\subsection{Parallel Feature Encoding for Motion and Object Saliency Estimation}
To encode the motion-salient regions in the current frame, we compute optical flow at cascaded positions along the temporal dimension. The optical flow between the current and previous frames at multiple distances are computed. We compute the optical flow map between the current frame and frames from temporal history with distance 1, 3 and 5, respectively. If the cascaded optical flow and the current frame are denoted with $C_{OF}$ and $I$, then the assimilated feature map is computed as given in Eq. \ref{eq1}.
\begin{equation}
AsOF=[C_{OF},I] \label{eq1}
\end{equation}
It provides crucial encoding to construct coarse motion saliency maps. We then extract deep features from $AsOF$ for higher-level abstractions. The features are extracted from three layers $(C3, C4, C5)$ of the ResNet residual stage 
as in~\cite{linfocalpami,lin2017feature,he2016deep}. We use resnet50 pre-trained over the ImageNet dataset as the backbone model for feature extraction. However, any other model can also be used as a backbone model. Moreover, in order to reinforce the semantic features of the salient foreground/moving objects more accurately, backbone features are also parallelly extracted from the current frame. These two sets of backbone feature maps are combined at matching scales for both temporal and spatial saliency aware feature representation. The motion-salient features are encoded as given in Eq.~\ref{eq2}.
\begin{equation}
MSF=[resnet(AsOF),resnet(I)] \label{eq2}
\end{equation}
where $resnet(x)$ returns the features from ResNet50 backbone.

\begin{figure}[]
\centering
	\includegraphics[width=0.5\textwidth ]{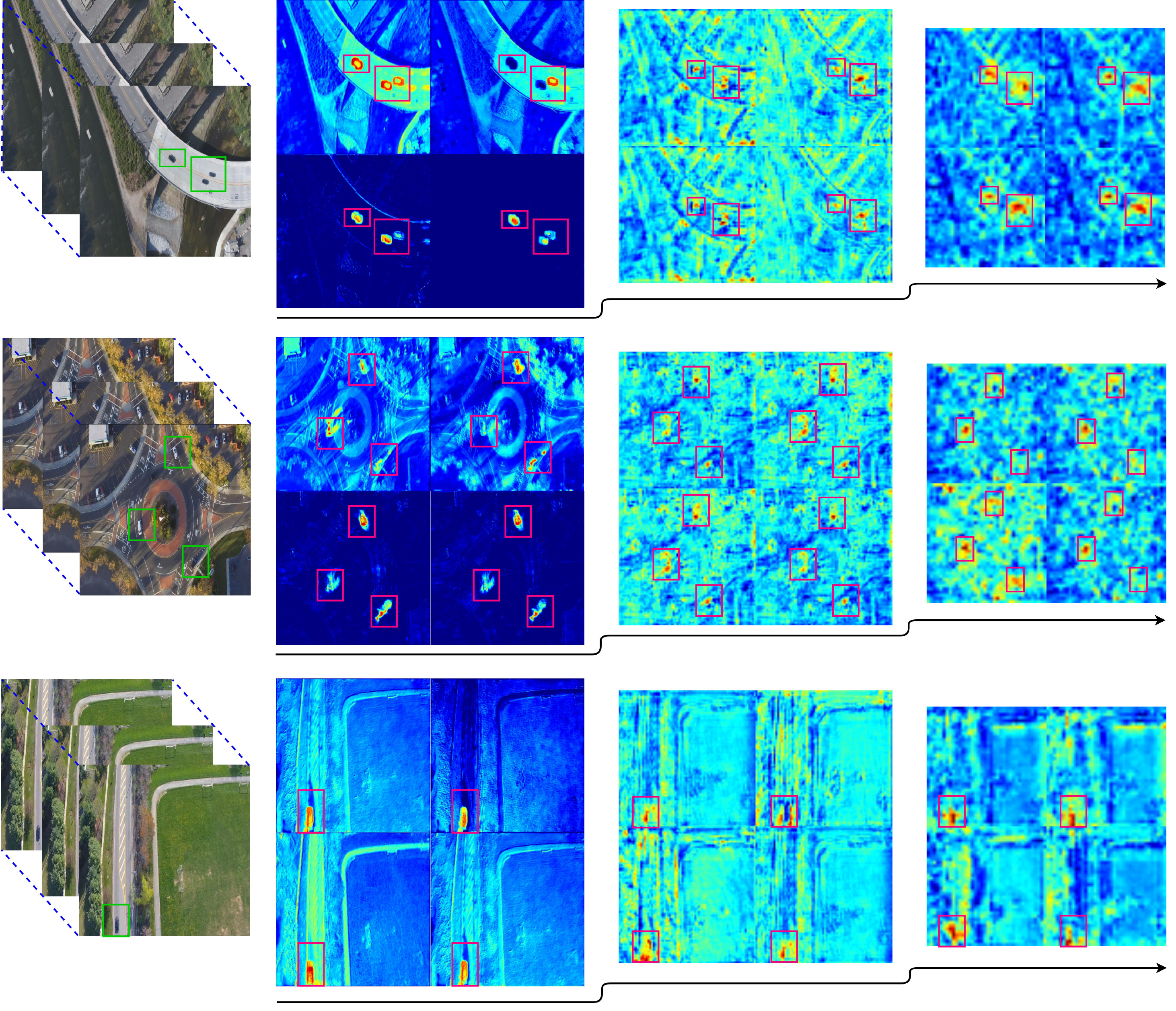}
	\hfill\small \center \hspace{2cm}(a) \hspace{2cm} (b) \hspace{2cm} (c) \caption{Visualization of different layers in MOR-UAVNet. (a) convolutional layer before the backbone feature extraction, (b) pyramid level–P3, (c) pyramid level–P4. The relevant motion saliencies of moving objects are highlighted using red boxes.}
	\label{fig:figure5}
\end{figure}


\subsection{Baseline Models}
Since this is the first attempt for MOR in UAV videos, we compute the baseline results using the proposed MOR-UAVNet framework. In addition to the proposed MOR-UAVNetv1, we also designed MOR-UAVNetv2 by removing the parallel feature extraction (computed separately for the current frame) part from the original network. Thus, we could also evaluate the effect of directly using AsOF without the reinforcements of base features extracted from the current frame.  We created MOR-UAVNetv1, MOR-UAVNetv2 by using resnet50~\cite{he2016deep} and MOR-UAVNetv3, MOR-UAVNetv4 by using mobileNetv2~\cite{sandler2018mobilenetv2} as backbone respectively. Thus, we have four different networks to compute baseline results on MOR-UAV dataset. Each network is trained four times for $T=1 (C_{OF}=1), T=2 (C_{OF}=1,3), T=2 (C_{OF}=1,5)$ and $T=3 (C_{OF}=1,3,5)$, respectively. Thus, overall, we evaluate the quantitative results for 16 models over MOR-UAV dataset.

\subsection{Training, Inference \& Visualization}
\textbf{Network Configurations.} We resize all the video frames in MOR-UAV dataset to $608\times 608\times 3$ for a uniform setting in training and evaluation. The MOR-UAVNet takes two tensors of shape $608\times 608\times T$ (number of cascaded optical flow maps) and $608\times 608\times 3$ (current frame) as input and returns the spatial coordinates with class labels for moving object instances. We compute the dense optical flow using the algorithm given in~\cite{farneback2003two}. The following values of $T$ is used in our experiments: $T=3 (C_{OF}=1-3-5), T=2 (C_{OF}=1-3), T=2 (C_{OF}=1-5) and T=1 (C_{OF}=1)$ in our experiments.\par

\textbf{Training.} The one-stage MOR-UAVNet network is trained end-to-end with multiple input layers. The complete framework is implemented in Keras with Tensorflow backend. Training is performed with $batch size=1$ over 11 GB Nvidia RTX 2080 Ti GPU. Due to the memory limitation and large image-size, we use $batch size=1$ in our experiments. The network is optimized with Adam optimizer and initial learning rate of $10^{-5}$. All models are trained for approximately $~250K-300K$ iterations. For regression and classification, $smooth \ell_1$ and focal loss functions are used respectively. The training loss is computed as the sum of the above-mentioned two losses. \par

\textbf{Inference.} Similar to training, inference involves simply giving the current frame and $T$ cascaded optical flow maps computed from past history frames as input to the network. Only a few optical flow maps $(T=1/2/3)$ are required, enabling online moving object recognition for real-time analysis. Top $1000$ prediction scores per pyramid level are considered after thresholding detector confidence at $0.05$. The final detections are collected by combining top predictions from all levels and non-maximum suppression with a threshold of $0.5$. \par

\textbf{Visualization.} We depict the intermediate layer visualizations of MOR-UAVNet in Figure~\ref{fig:figure5}. Here, we can see that the salient-motion regions are robustly localized from recent temporal history. The pyramid levels are able to represent the moving objects at multiple scales. Further processing through detection and classification sub-networks results in accurate MOR.

\begin{table}[t]
\centering
\caption{Summary description of the training and testing sets used in our experiments for evaluation}
\begin{tabular}{c c c c c} 
\hline 
 Videos & \#Frames & \#Car & \begin{tabular}[c]{@{}l@{}}\#Heavy\\ Vehicle\end{tabular} & \#Objects\\
 \hline 
video1 & 299 & 6,026 & 600 & 6,626\\ 
video2 & 499 & 6,584 & 493 & 7,077\\ 
video3 & 225 & 1,575 & 0 & 1,575\\ 
video4 & 139 & 695 & 138 & 833\\ 
video5 & 351 & 1,049 & 610 & 1,659\\ 
video6 & 218 & 1,902 & 156 & 2,058\\ 
video7 & 64 & 65 & 42 & 107\\ 
video8 & 118 & 187 & 0 & 187\\ video9 & 299 & 4,571 & 251 & 4,822\\ 
video10 & 477 & 12,579 & 2,527 & 15,106\\ 
video11 & 225 & 8,699 & 2,617 & 11,316\\ 
video12 & 550 & 940 & 78 & 1,018\\ 
video13 & 285 & 0 & 286 & 286\\ 
video13 & 285 & 0 & 286 & 286\\ 
\hline
\textit{Training Set} & \textit{3,749} & \textit{44,872} & \textit{7,798} & \textit{52,670}\\ 
\hline
video14 & 210 & 430 & 0 & 430\\ 
video15 & 199 & 626 & 0 & 626\\ 
video16 & 200 & 457 & 0 & 457\\ 
video17 & 70 & 427 & 71 & 498\\ 
\hline
\textit{Testing Set} & \textit{679} & \textit{1,940} & \textit{71} & \textit{2,011}\\ 
\hline
\end{tabular}
\label{table2}
\end{table}

\section{Experiments and Discussions}
In this section, we present the training and testing data description, evaluation metrics and the baseline results computed with the proposed MOR-UAVNet framework. We give a detailed quantitative and qualitative analysis of the proposed baseline methods to report the effectiveness of our novel MOR framework. 

\textbf{Dataset.} From the MOR-UAV dataset, we select a subset of videos for training and evaluation. The training set consists of $13$ video sequences having $3,749$ frames and $52,670$ objects ($44,872$ cars and $7,798$ heavy vehicles). The testing set consists of 4 video sequences with $679$ frames and $2,011$ objects ($1,940$ cars and $71$ heavy vehicles). The detailed description of train and test sets used for qualitative and quantitative evaluation is given in Table~\ref{table2}. The proposed labels can also be used to design and evaluate class-agnostic moving object detection algorithms.

\textbf{Evaluation.} As the results are computed in terms of spatial coordinates and class labels of moving object instances in every frame. Thus, to measure the MOR performance, we use the standard average precision (AP)~\cite{linfocalpami} metrics. Evaluation is performed with IoU threshold $0.50$, i.e. predicted an object instance is considered true positive if it has at least $50\%$ intersection-over-union (IoU) with the corresponding ground truth object instance. The mean average precision $mAP50$ computes the means of APs across two classes: car and heavy vehicle. It is to be noted that the mAP is computed only for object instances with movements. The remaining objects (non-moving) are part of the background according to the definition of MOR.

\subsection{Performance Analysis}
\begin{figure*}[]
\centering
\begin{subfigure}{.3\linewidth}
    \centering
    \includegraphics[width=1\textwidth]{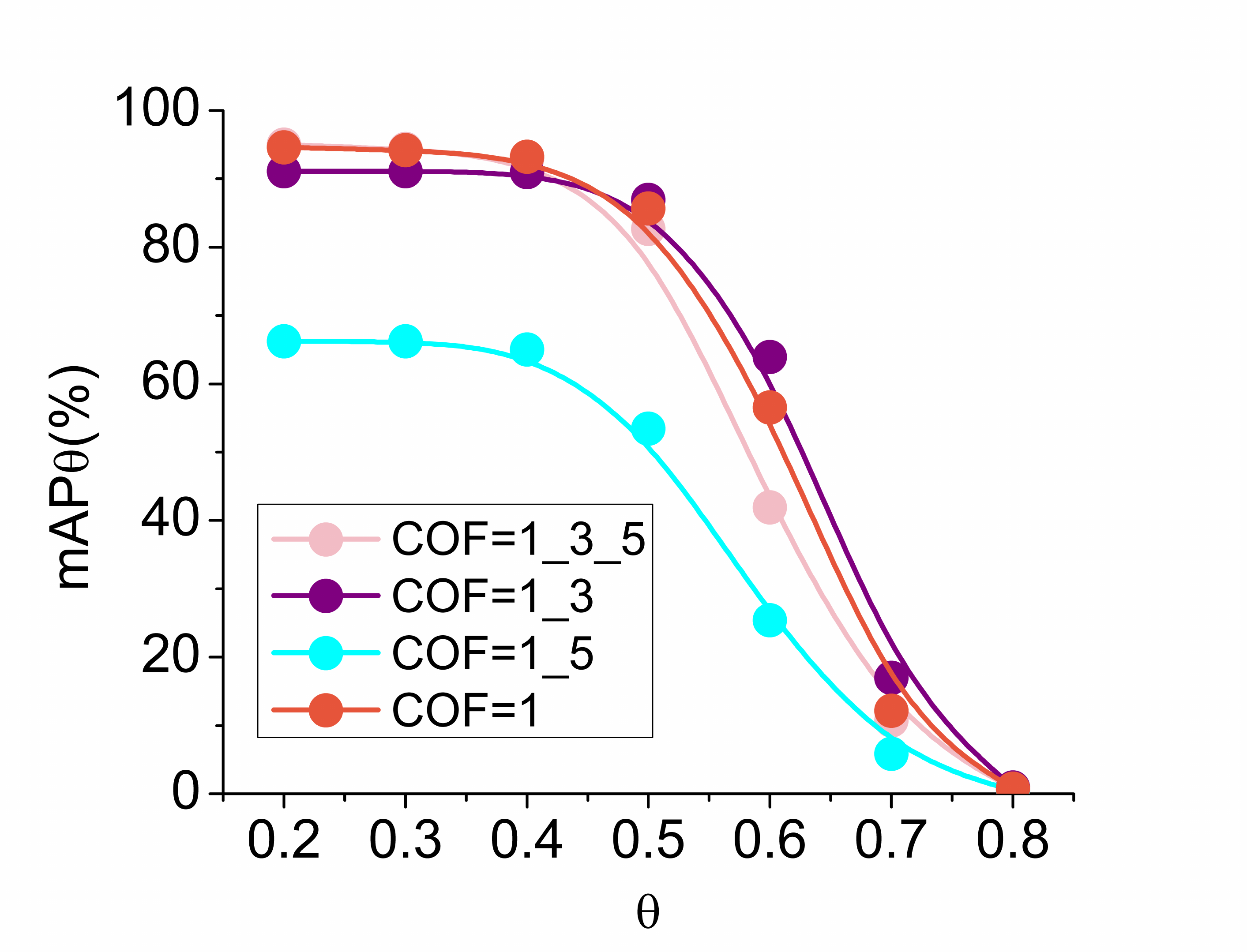}
    \caption{}\label{fig:6a}
\end{subfigure}
    \hfill
\begin{subfigure}{.3\linewidth}
    \centering
    \includegraphics[width=1\textwidth]{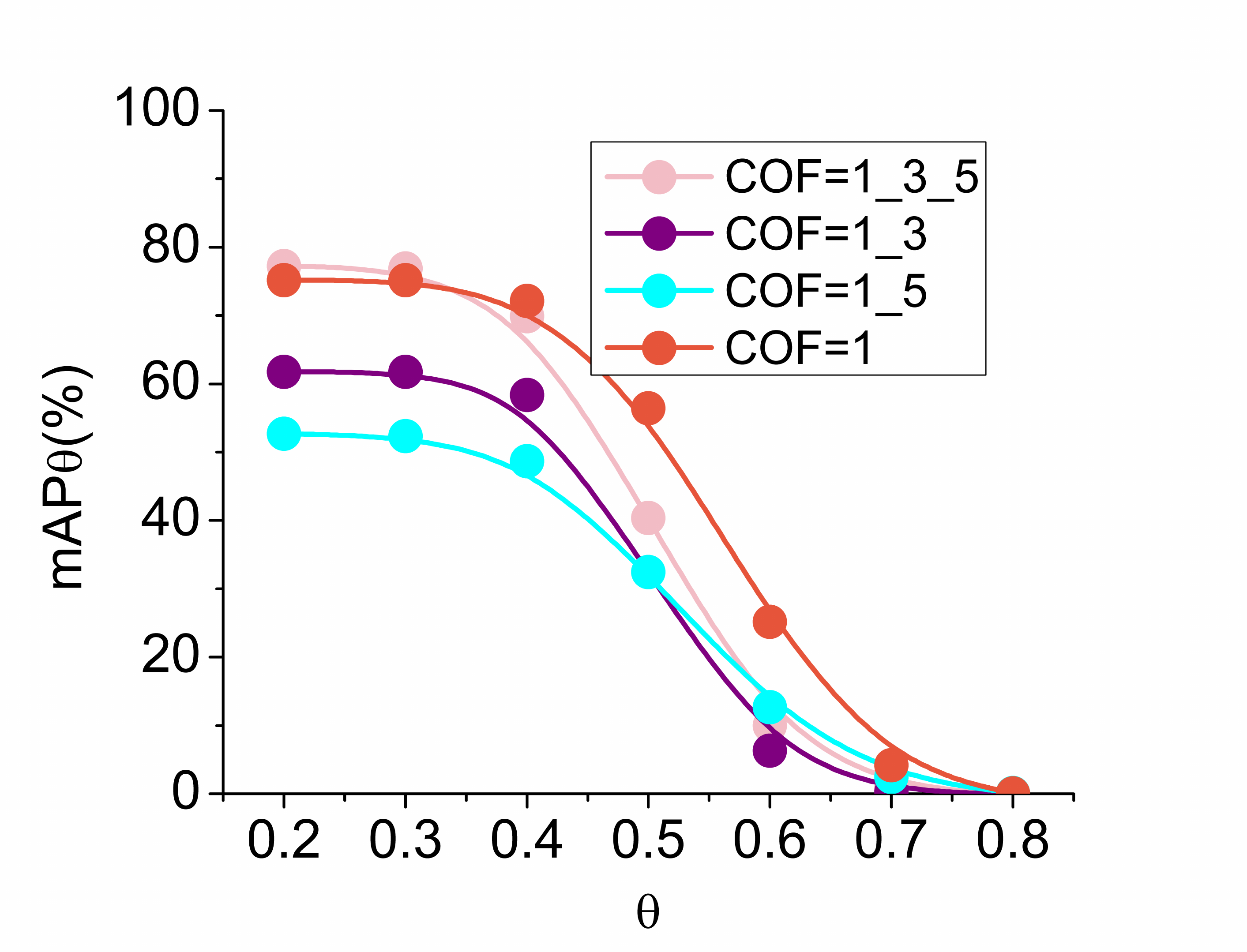}
    \caption{}\label{fig:6b}
\end{subfigure}
   \hfill
\begin{subfigure}{.3\linewidth}
    \centering
    \includegraphics[width=1\textwidth]{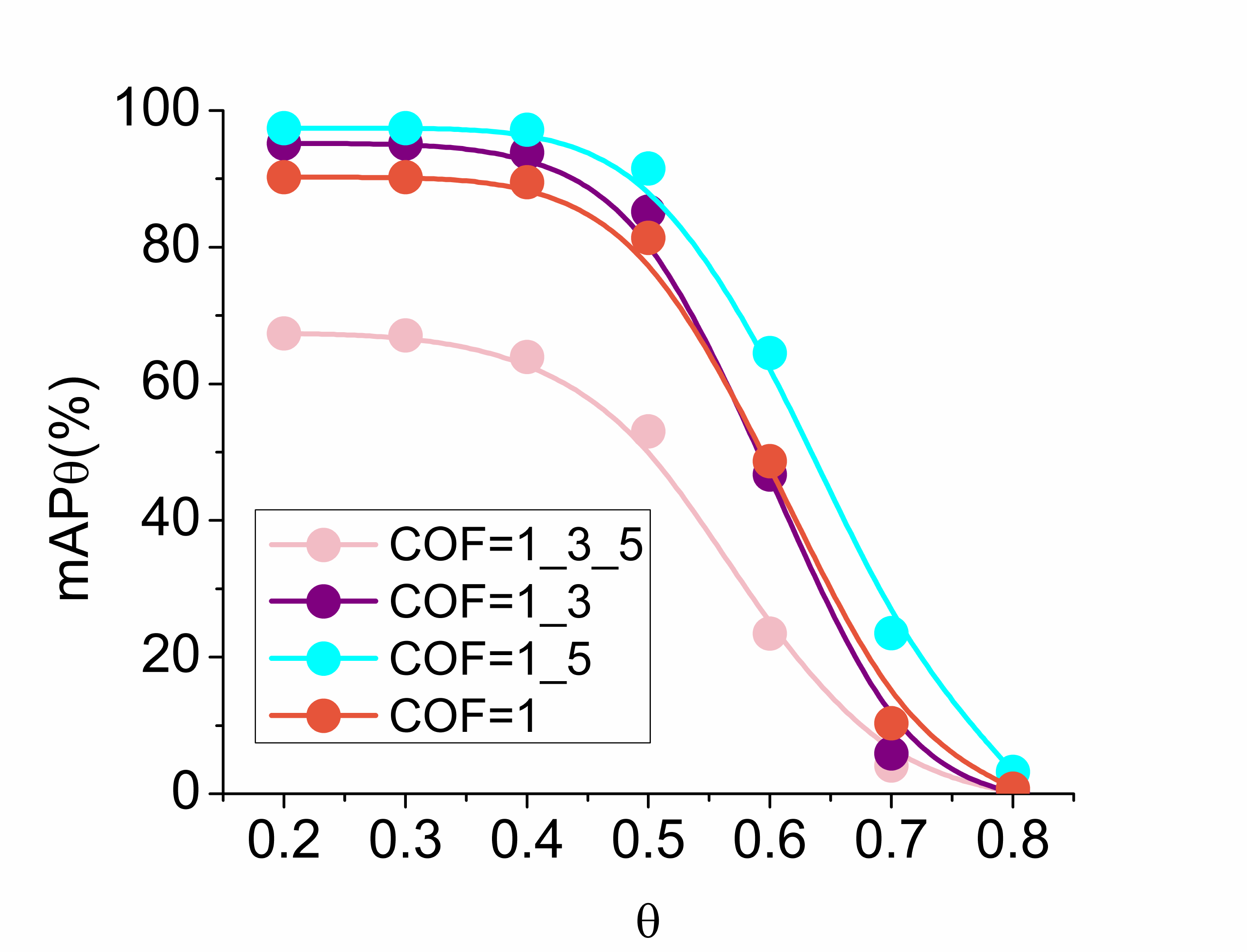}
    \caption{}\label{fig:6c}
\end{subfigure}

\bigskip
\begin{subfigure}{0.4\linewidth}
  \centering
  \includegraphics[width=0.7\textwidth]{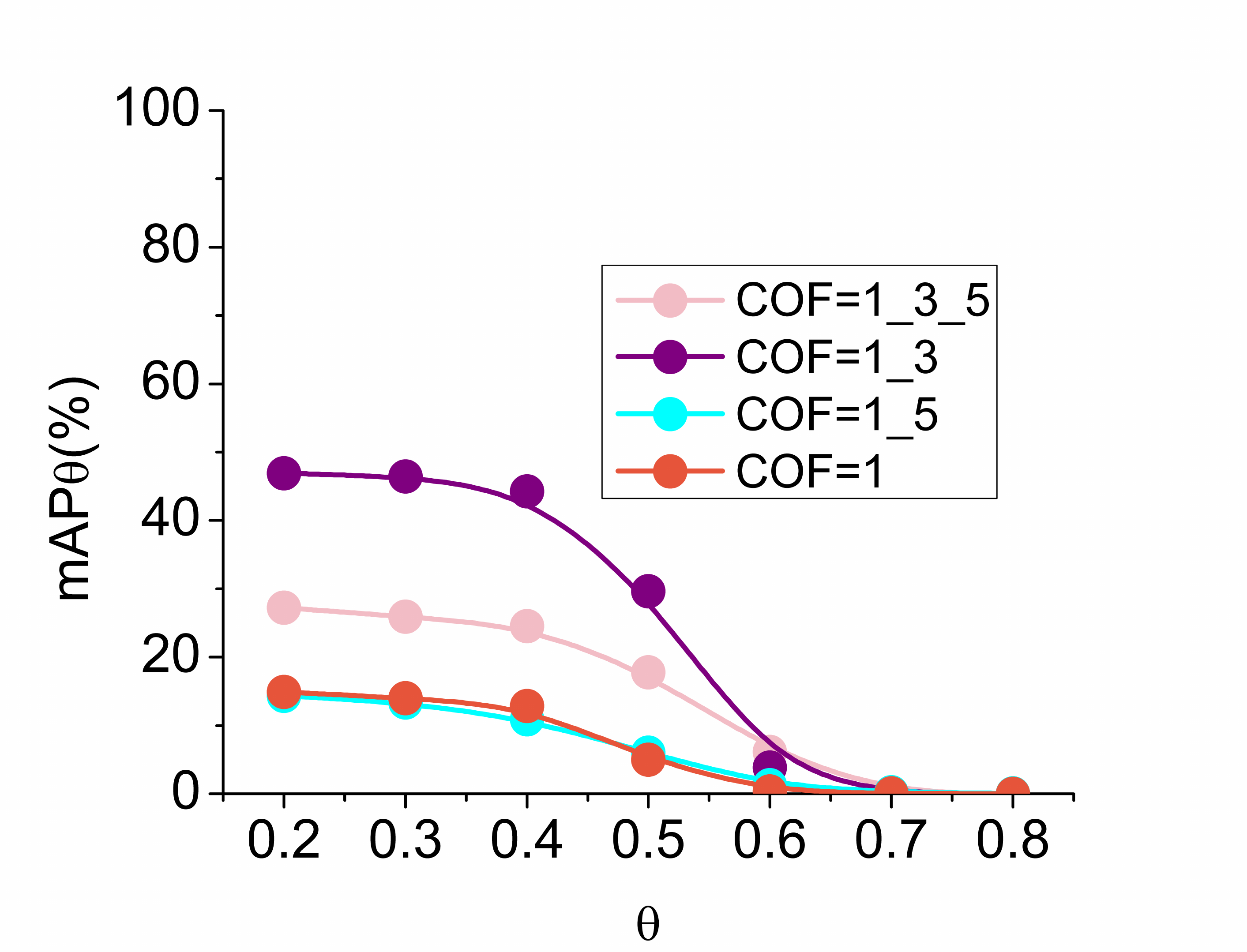}
  \caption{}\label{fig:6d}
\end{subfigure} 
\begin{subfigure}{0.4\linewidth}
    \centering
    \includegraphics[width=0.7\textwidth]{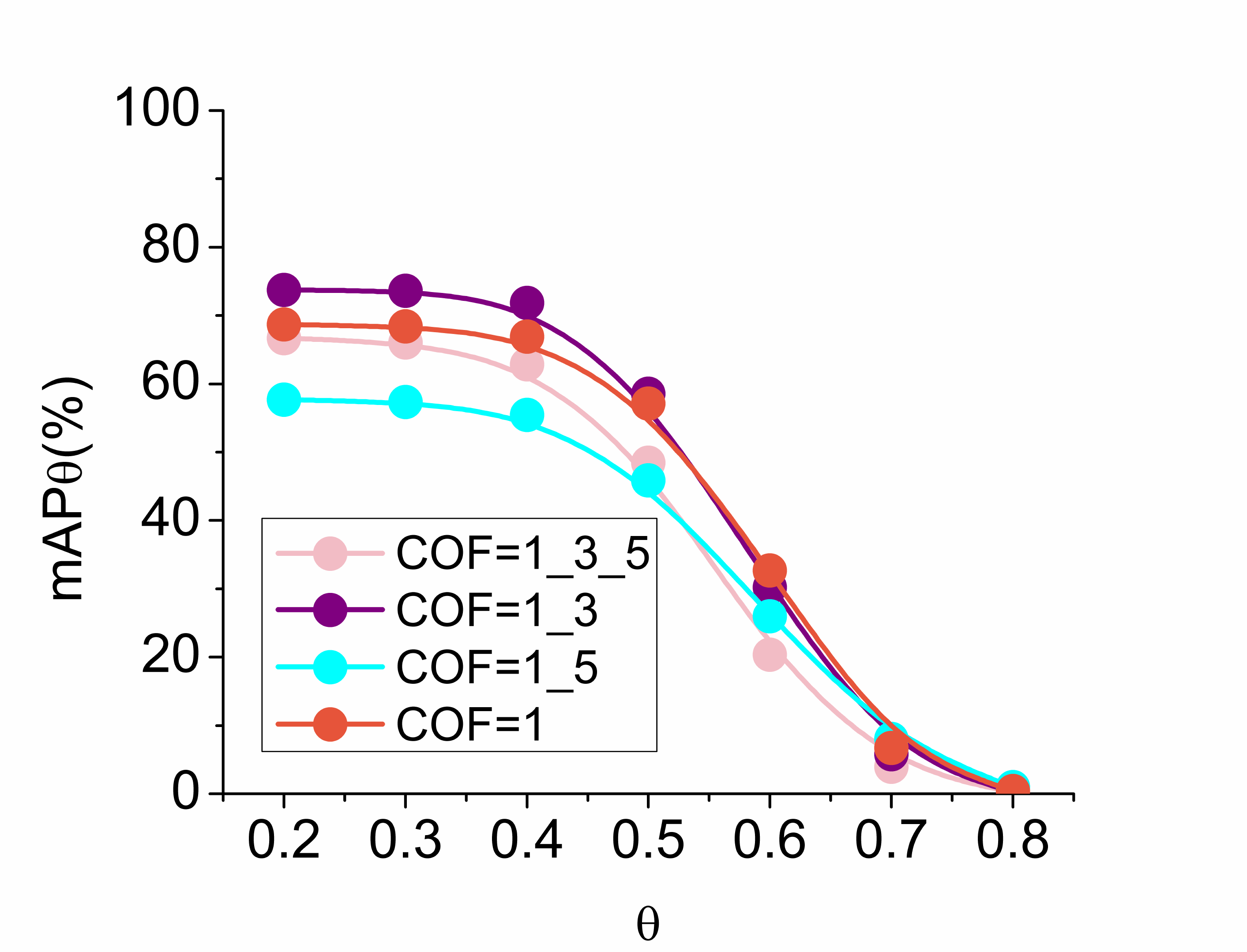}
    \caption{}\label{fig:6e}
\end{subfigure}
\caption{MOR mAP of MOR-UAVNetv1 across different IoU thresholds over (a) vid14, (b) vid15, (c) vid16, (d) vid17 videos and (e) average across all video sequences. $C_{OF}$ is the cascaded optical flow maps used in the input layer}
\label{fig:figure6}
\end{figure*}

\textbf{Quantitative analysis.} From Table~\ref{table3}, it can be noticed that MOR-UAVNetv1 performs better than MOR-UAVNetv2 in vid14, vid15, and vid16. The MOR-UAVNetv2 performs better in vid17. However, MOR-UAVNetv1 outperforms MOR-UAVNetv2 in overall results. The models MOR-UAVNetv3, MOR-UAVNetv4 with mobileNetv2 backbone also perform reasonably well and can be considered for the resource-constrained environment. The best performing model achieves mAP of $58.56$ which highlights the challenging nature of the MOR-UAV dataset. The mAP at different IoUs for MOR-UAVNetv1 is further analyzed through Figure~\ref{fig:figure6}. The IoU vs mAP graph for MOR across different $C_{OF}$ depths for each test video is depicted in Figure~\ref{fig:figure6} (a), Figure~\ref{fig:figure6} (b), Figure~\ref{fig:figure6}(c) and Figure~\ref{fig:figure6} (d) respectively. The mAP is computed at IoU thresholds $0.2, 0.3, 0.4, 0.5, 0.6, 0.7, 0.8$. We also show the overall average performance through the graph in Figure~\ref{fig:figure6} (e). It is clear that if we could lower the IoU threshold, we can recognize a higher number of moving objects. However, it might also increase the number of false detections. The decision for the same can be taken according to the demands of real-world applications.\par

\begin{table}[]
\centering
\caption{Quantitative results (mAP) of the proposed MOR-UAVNet based baseline models over the MOR-UAV dataset. The best results are highlighted in \textbf{bold}}
\begin{tabular}{c|c c c c c c} 
 \hline 
 Method & $C_{OF}$ & Vid14 & Vid15 & Vid16 & Vid17 & Avg\\
 \hline
& 1-3-5 &82.70&	40.35&	53.06&	17.76&	48.47\\
MOR-& 1-3 & \textbf{86.94}&	32.48&	85.18&	29.64&	\textbf{58.56}\\
UAVNetv1& 1-5 & 53.43&	32.45&	\textbf{91.49}&	6.04&	45.85\\
& 1 & 85.65&	56.41&	81.35&	4.98&	57.09\\
\hline

 & 1-3-5 & 71.12&	19.68&	69.46&	29.79&	47.51\\
MOR-& 1-3 & 79.31&	40.51&	59.06&	31.75&	52.65\\
UAVNetv2& 1-5 & 83.18&	38.02&	80.53&	30.66&	58.09\\
& 1 & 
85.57&	23.34&	19.17&	39.02&	41.77\\
\hline

 & 1-3-5 & 
39.04&	35.73&	16.59&	\textbf{49.14}&	35.13\\
MOR-& 1-3 & 60.54&	25.91&	05.25&	17.54&	27.31\\
UAVNetv3& 1-5 & 71.61&	33.94&	61.46&	8.06&	43.77\\
& 1 & 79.04&	44.09&	72.85&	19.27&	53.81\\
\hline

 & 1-3-5 & 
60.90&	32.65&	79.52&	14.72&	46.95\\
MOR-& 1-3 & 65.41&	48.82&	38.28&	29.01&	45.38\\
UAVNetv4& 1-5 & 80.59&	41.14&	62.47&	19.48&	50.92\\
& 1 & 58.47&	\textbf{57.62}&	41.67&	5.71&	40.87\\
\hline
\end{tabular}
\label{table3}
\end{table}

\begin{table}[]
\small
\centering
\caption{Inference speed, number of parameters and model size comparison of the baseline models.}
\begin{tabular}{c|c c c c} 
 \hline 
 Method & $C_{OF}$ & FPS & \#Param & Model Size\\
 \hline
 
\multirow{3}{*}{MOR-UAVNetv1} & 1-3-5 & 9.59 & \multirow{3}{*}{{\raise.17ex\hbox{$\scriptstyle\sim$}}65.4M} & \multirow{3}{*}{{\raise.17ex\hbox{$\scriptstyle\sim$}}263.6MB}\\ 
& 1-3/1-5 & 9.59 & &\\
& 1 & 11.11 & &\\
\hline
\multirow{3}{*}{MOR-UAVNetv2} & 1-3-5 & 8.44 & \multirow{3}{*}{{\raise.17ex\hbox{$\scriptstyle\sim$}}36.3M} & \multirow{3}{*}{{\raise.17ex\hbox{$\scriptstyle\sim$}}146.3MB}\\ 
& 1-3/1-5 & 8.79 & &\\
& 1 & 10.05 & &\\
\hline
\multirow{3}{*}{MOR-UAVNetv3} & 1-3-5 & 7.81 & \multirow{3}{*}{{\raise.17ex\hbox{$\scriptstyle\sim$}}19.3M} & \multirow{3}{*}{{\raise.17ex\hbox{$\scriptstyle\sim$}}77.8MB}\\ 
& 1-3/1-5 & 8.79 & &\\
& 1 & 9.59 & &\\
\hline
\multirow{3}{*}{MOR-UAVNetv4} & 1-3-5 & 9.17 & \multirow{3}{*}{{\raise.17ex\hbox{$\scriptstyle\sim$}}13.2M} & \multirow{3}{*}{{\raise.17ex\hbox{$\scriptstyle\sim$}}53.3MB}\\ 
& 1-3/1-5 & 9.59 & &\\
& 1 & 10.55 & &\\
\hline
\end{tabular}
\label{table4}
\end{table}

\begin{figure*}[]
\centering
	\includegraphics[width=0.9\textwidth ]{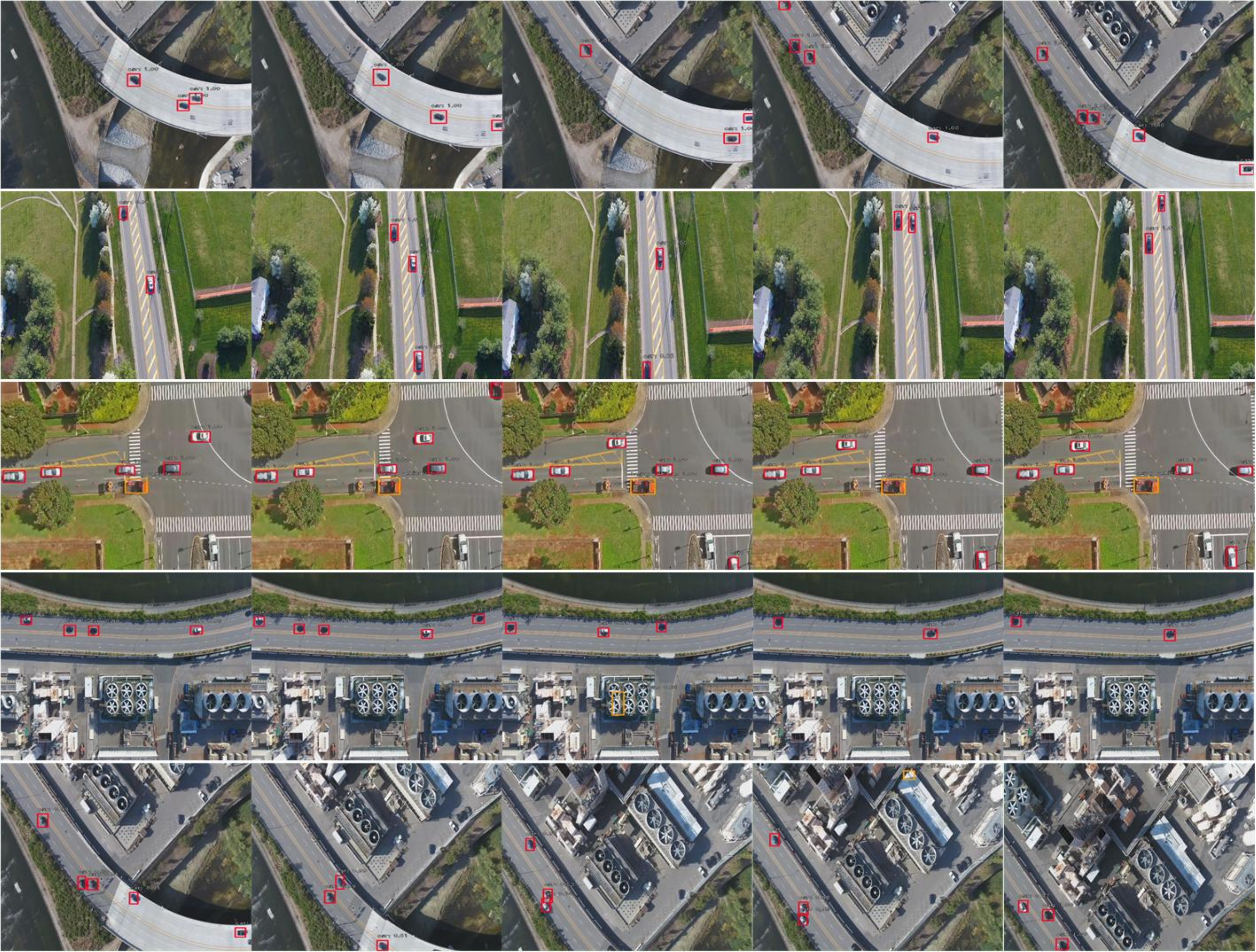}
	\caption{Qualitative results of MOR-UAVNetv1 over completely unseen video sequences in MOR-UAV dataset.}	\label{figure7}
\end{figure*}

\textbf{Qualitative analysis.} We show the qualitative results of the proposed method on five completely unseen videos in Figure~\ref{figure7}. The MOR-UAVNetv1 obtains reasonable performance in diverse objects and camera movements. For example, in the first, second and fifth rows, in addition to the object movements, the UAV camera itself is moving. Similarly, in the fourth row, the vehicles are moving nearby an industrial area with complex structures. In the third row, the moving and non-moving objects are adjacently located at some point. All these scenarios are handled quite well.\par

\begin{figure}[]
\centering
	\includegraphics[width=0.45\textwidth ]{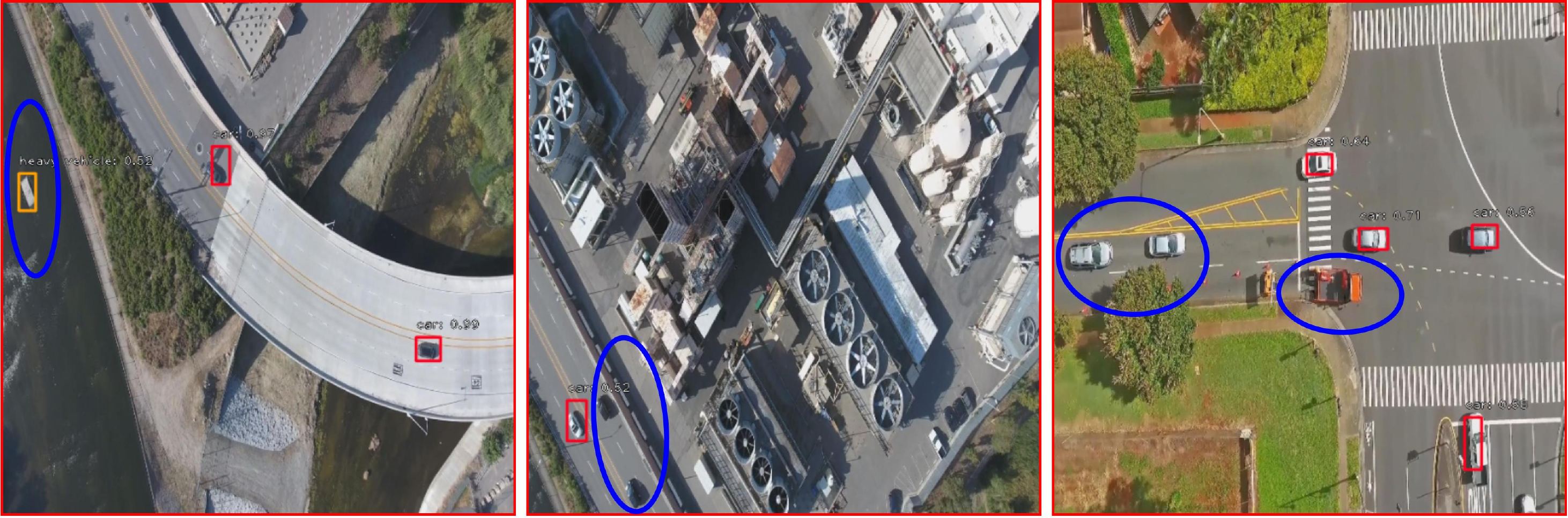}
	\hfill \small \center (a)\hspace{2cm}(b)\hspace{2cm}(c)
	\caption{Failure cases in MOR prediction with the MOR-UAVNetv1 model.} 
	\label{fig:figure8}
\end{figure}

\textbf{Run-time analysis.} We also tabulate the inference speed, compute and memory requirements of the proposed  methods in Table~\ref{table4}. The MOR-UAVNetv4 has the lowest number of trainable parameters and model size as compared to the other three versions. It also demonstrates the fastest inference at $\scriptstyle\sim11$ FPS in RTX 2080Ti which is quite reasonable. Since MOR-UAVNetv1 achieves better mAP, overall, it outperforms the remaining 3 models when all the performance measures are taken into consideration. However, more work needs to be done to develop faster and more accurate MOR algorithms for real-time applications.\par

\textbf{Failure cases.} We discuss three scenarios in which MOR-UAVNet fails as shown in Figure~\ref{fig:figure8}. The first case (Figure~\ref{fig:figure8}(a)) arises due to motion in a background object which does not belong to car or heavy vehicle category. In the second case (Figure~\ref{fig:figure8} (b)), the UAV camera is moving in the same direction as the objects making it difficult for the algorithm to identify the object motion accurately. In the last case (Figure~\ref{fig:figure8}(c)), the slowly moving objects are sometimes not detected by the proposed method. These failure cases can be overcome by including even more diversified scenarios in the training dataset and develop better algorithms for robust MOR performance in the future.

\subsection{Discussions}
Our benchmark caters to real-world demands with vivid samples collected from numerous unconstrained circumstances. Although the proposed MOR-UAVNet algorithms perform reasonably well on the test set, there is a lot of scopes to further improve the performance. We feel this benchmark dataset can support promising research trends in vehicular technology. We mention some of the future research directions for exploration.

\textbf{Realtime challenges.} Inference speed is one of the essential requirement for practical UAV based applications. The commonly deployed UAVs for aerial scene analysis are highly resource-constrained in nature. Although the proposed method achieves $\scriptstyle\sim11$ fps, even better speed is desired. Thus, more MOR algorithms are needed that can robustly operate with better accuracy and speed in resource-constrained environment. Some recent works~\cite{tan2019mnasnet,wu2019fbnet} have shown promising directions to increase efficiency through neural architectural search, network pruning  and compression. We expect future works to develop solutions to address both accuracy and real-time constraints.

\textbf{Locating motion clues.} The motion appearance varies among different video sequences in the dataset. For example, variable speed (fast, medium, slow) of moving vehicles, unconstrained camera motion and rotation resulting in changing backgrounds make it difficult to accurately identify the motion clues. Moreover, the object scales also fluctuate between small, medium and large shapes due to changes in the altitude of UAV device. Thus, to obtain robust performance, these demanding factors need to be addressed in future methods.

\section{Conclusions}
This paper introduces a challenging unconstrained benchmark MOR-UAV dataset for the task of moving object recognition (MOR) in UAV videos.
A deep learning framework MOR-UAVNet is presented along with 16 different models for baseline results. As a first largescale labeled dataset dedicated to MOR in UAV videos, the proposed work contributes to vehicular vision community by establishing a new benchmark. The  proposed MOR-UAVNet framework may also be used to design new algorithms to advance MOR research in the future.

\section{Acknowledgements}
The authors are highly grateful to IBM for providing with online GPU grant. The work was also supported by the DST-SERB project \#EEQ/2017/000673.

\bibliographystyle{ACM-Reference-Format}
\bibliography{main.bib}


\begin{thebibliography}{75}


\ifx \showCODEN    \undefined \def \showCODEN     #1{\unskip}     \fi
\ifx \showDOI      \undefined \def \showDOI       #1{#1}\fi
\ifx \showISBNx    \undefined \def \showISBNx     #1{\unskip}     \fi
\ifx \showISBNxiii \undefined \def \showISBNxiii  #1{\unskip}     \fi
\ifx \showISSN     \undefined \def \showISSN      #1{\unskip}     \fi
\ifx \showLCCN     \undefined \def \showLCCN      #1{\unskip}     \fi
\ifx \shownote     \undefined \def \shownote      #1{#1}          \fi
\ifx \showarticletitle \undefined \def \showarticletitle #1{#1}   \fi
\ifx \showURL      \undefined \def \showURL       {\relax}        \fi
\providecommand\bibfield[2]{#2}
\providecommand\bibinfo[2]{#2}
\providecommand\natexlab[1]{#1}
\providecommand\showeprint[2][]{arXiv:#2}

\bibitem[\protect\citeauthoryear{Akilan and Wu}{Akilan and Wu}{2019}]%
        {akilan2019sendec}
\bibfield{author}{\bibinfo{person}{Thangarajah Akilan} {and}
  \bibinfo{person}{QM~Jonathan Wu}.} \bibinfo{year}{2019}\natexlab{}.
\newblock \showarticletitle{sEnDec: An improved image to image CNN for
  foreground localization}.
\newblock \bibinfo{journal}{\emph{IEEE Transactions on Intelligent
  Transportation Systems}} (\bibinfo{year}{2019}).
\newblock


\bibitem[\protect\citeauthoryear{An, Choi, Kim, and Kim}{An
  et~al\mbox{.}}{2019}]%
        {an2019new}
\bibfield{author}{\bibinfo{person}{Jhonghyun An}, \bibinfo{person}{Baehoon
  Choi}, \bibinfo{person}{Hyunju Kim}, {and} \bibinfo{person}{Euntai Kim}.}
  \bibinfo{year}{2019}\natexlab{}.
\newblock \showarticletitle{A New Contour-Based Approach to Moving Object
  Detection and Tracking Using a Low-End Three-Dimensional Laser Scanner}.
\newblock \bibinfo{journal}{\emph{IEEE Transactions on Vehicular Technology}}
  \bibinfo{volume}{68}, \bibinfo{number}{8} (\bibinfo{year}{2019}),
  \bibinfo{pages}{7392--7405}.
\newblock


\bibitem[\protect\citeauthoryear{Bao, Wu, and Liu}{Bao et~al\mbox{.}}{2018}]%
        {bao2018cnn}
\bibfield{author}{\bibinfo{person}{Linchao Bao}, \bibinfo{person}{Baoyuan Wu},
  {and} \bibinfo{person}{Wei Liu}.} \bibinfo{year}{2018}\natexlab{}.
\newblock \showarticletitle{CNN in MRF: Video object segmentation via inference
  in a CNN-based higher-order spatio-temporal MRF}. In
  \bibinfo{booktitle}{\emph{Proceedings of the IEEE Conference on Computer
  Vision and Pattern Recognition}}. \bibinfo{pages}{5977--5986}.
\newblock


\bibitem[\protect\citeauthoryear{Barekatain, Mart{\'\i}, Shih, Murray,
  Nakayama, Matsuo, and Prendinger}{Barekatain et~al\mbox{.}}{2017}]%
        {barekatain2017okutama}
\bibfield{author}{\bibinfo{person}{Mohammadamin Barekatain},
  \bibinfo{person}{Miquel Mart{\'\i}}, \bibinfo{person}{Hsueh-Fu Shih},
  \bibinfo{person}{Samuel Murray}, \bibinfo{person}{Kotaro Nakayama},
  \bibinfo{person}{Yutaka Matsuo}, {and} \bibinfo{person}{Helmut Prendinger}.}
  \bibinfo{year}{2017}\natexlab{}.
\newblock \showarticletitle{Okutama-action: An aerial view video dataset for
  concurrent human action detection}. In \bibinfo{booktitle}{\emph{Proceedings
  of the IEEE Conference on Computer Vision and Pattern Recognition
  Workshops}}. \bibinfo{pages}{28--35}.
\newblock


\bibitem[\protect\citeauthoryear{Bazi and Melgani}{Bazi and Melgani}{2018}]%
        {bazi2018convolutional}
\bibfield{author}{\bibinfo{person}{Yakoub Bazi} {and} \bibinfo{person}{Farid
  Melgani}.} \bibinfo{year}{2018}\natexlab{}.
\newblock \showarticletitle{Convolutional SVM networks for object detection in
  UAV imagery}.
\newblock \bibinfo{journal}{\emph{Ieee transactions on geoscience and remote
  sensing}} \bibinfo{volume}{56}, \bibinfo{number}{6} (\bibinfo{year}{2018}),
  \bibinfo{pages}{3107--3118}.
\newblock


\bibitem[\protect\citeauthoryear{Berker~Logoglu, Lezki, Kerim~Yucel, Ozturk,
  Kucukkomurler, Karagoz, Erdem, and Erdem}{Berker~Logoglu
  et~al\mbox{.}}{2017}]%
        {berker2017feature}
\bibfield{author}{\bibinfo{person}{K Berker~Logoglu}, \bibinfo{person}{Hazal
  Lezki}, \bibinfo{person}{M Kerim~Yucel}, \bibinfo{person}{Ahu Ozturk},
  \bibinfo{person}{Alper Kucukkomurler}, \bibinfo{person}{Batuhan Karagoz},
  \bibinfo{person}{Erkut Erdem}, {and} \bibinfo{person}{Aykut Erdem}.}
  \bibinfo{year}{2017}\natexlab{}.
\newblock \showarticletitle{Feature-based efficient moving object detection for
  low-altitude aerial platforms}. In \bibinfo{booktitle}{\emph{Proceedings of
  the IEEE International Conference on Computer Vision Workshops}}.
  \bibinfo{pages}{2119--2128}.
\newblock


\bibitem[\protect\citeauthoryear{Chen, Zhu, Papandreou, Schroff, and Adam}{Chen
  et~al\mbox{.}}{2018}]%
        {chen2018encoder}
\bibfield{author}{\bibinfo{person}{Liang-Chieh Chen}, \bibinfo{person}{Yukun
  Zhu}, \bibinfo{person}{George Papandreou}, \bibinfo{person}{Florian Schroff},
  {and} \bibinfo{person}{Hartwig Adam}.} \bibinfo{year}{2018}\natexlab{}.
\newblock \showarticletitle{Encoder-decoder with atrous separable convolution
  for semantic image segmentation}. In \bibinfo{booktitle}{\emph{Proceedings of
  the European conference on computer vision (ECCV)}}.
  \bibinfo{pages}{801--818}.
\newblock


\bibitem[\protect\citeauthoryear{Cheng, Tsai, Wang, and Yang}{Cheng
  et~al\mbox{.}}{2017}]%
        {cheng2017segflow}
\bibfield{author}{\bibinfo{person}{Jingchun Cheng}, \bibinfo{person}{Yi-Hsuan
  Tsai}, \bibinfo{person}{Shengjin Wang}, {and} \bibinfo{person}{Ming-Hsuan
  Yang}.} \bibinfo{year}{2017}\natexlab{}.
\newblock \showarticletitle{Segflow: Joint learning for video object
  segmentation and optical flow}. In \bibinfo{booktitle}{\emph{Proceedings of
  the IEEE international conference on computer vision}}.
  \bibinfo{pages}{686--695}.
\newblock


\bibitem[\protect\citeauthoryear{Collins, Zhou, and Teh}{Collins
  et~al\mbox{.}}{2005}]%
        {collins2005open}
\bibfield{author}{\bibinfo{person}{Robert Collins}, \bibinfo{person}{Xuhui
  Zhou}, {and} \bibinfo{person}{Seng~Keat Teh}.}
  \bibinfo{year}{2005}\natexlab{}.
\newblock \showarticletitle{An open source tracking testbed and evaluation web
  site}. In \bibinfo{booktitle}{\emph{IEEE International Workshop on
  Performance Evaluation of Tracking and Surveillance}},
  Vol.~\bibinfo{volume}{35}.
\newblock


\bibitem[\protect\citeauthoryear{Deng, Dong, Socher, Li, Li, and Fei-Fei}{Deng
  et~al\mbox{.}}{2009}]%
        {deng2009imagenet}
\bibfield{author}{\bibinfo{person}{Jia Deng}, \bibinfo{person}{Wei Dong},
  \bibinfo{person}{Richard Socher}, \bibinfo{person}{Li-Jia Li},
  \bibinfo{person}{Kai Li}, {and} \bibinfo{person}{Li Fei-Fei}.}
  \bibinfo{year}{2009}\natexlab{}.
\newblock \showarticletitle{Imagenet: A large-scale hierarchical image
  database}. In \bibinfo{booktitle}{\emph{2009 IEEE conference on computer
  vision and pattern recognition}}. Ieee, \bibinfo{pages}{248--255}.
\newblock


\bibitem[\protect\citeauthoryear{Ding, Xue, Long, Xia, and Lu}{Ding
  et~al\mbox{.}}{2019}]%
        {ding2019learning}
\bibfield{author}{\bibinfo{person}{Jian Ding}, \bibinfo{person}{Nan Xue},
  \bibinfo{person}{Yang Long}, \bibinfo{person}{Gui-Song Xia}, {and}
  \bibinfo{person}{Qikai Lu}.} \bibinfo{year}{2019}\natexlab{}.
\newblock \showarticletitle{Learning roi transformer for oriented object
  detection in aerial images}. In \bibinfo{booktitle}{\emph{Proceedings of the
  IEEE Conference on Computer Vision and Pattern Recognition}}.
  \bibinfo{pages}{2849--2858}.
\newblock


\bibitem[\protect\citeauthoryear{Du, Zhu, Wen, Bian, Ling, Hu, Zheng, Peng,
  Wang, Zhang, et~al\mbox{.}}{Du et~al\mbox{.}}{2019}]%
        {du2019visdrone}
\bibfield{author}{\bibinfo{person}{Dawei Du}, \bibinfo{person}{Pengfei Zhu},
  \bibinfo{person}{Longyin Wen}, \bibinfo{person}{Xiao Bian},
  \bibinfo{person}{Haibin Ling}, \bibinfo{person}{Qinghua Hu},
  \bibinfo{person}{Jiayu Zheng}, \bibinfo{person}{Tao Peng},
  \bibinfo{person}{Xinyao Wang}, \bibinfo{person}{Yue Zhang}, {et~al\mbox{.}}}
  \bibinfo{year}{2019}\natexlab{}.
\newblock \showarticletitle{VisDrone-SOT2019: The Vision Meets Drone Single
  Object Tracking Challenge Results}. In \bibinfo{booktitle}{\emph{Proceedings
  of the IEEE International Conference on Computer Vision Workshops}}.
  \bibinfo{pages}{0--0}.
\newblock


\bibitem[\protect\citeauthoryear{Dwivedi, Mandal, Yadav, and Vipparthi}{Dwivedi
  et~al\mbox{.}}{2020}]%
        {dwivedi20203d}
\bibfield{author}{\bibinfo{person}{Shivangi Dwivedi}, \bibinfo{person}{Murari
  Mandal}, \bibinfo{person}{Shekhar Yadav}, {and}
  \bibinfo{person}{Santosh~Kumar Vipparthi}.} \bibinfo{year}{2020}\natexlab{}.
\newblock \showarticletitle{3D CNN with Localized Residual Connections for
  Hyperspectral Image Classification}. In \bibinfo{booktitle}{\emph{Computer
  Vision and Image Processing: 4th International Conference, CVIP 2019, Jaipur,
  India, September 27--29, 2019, Revised Selected Papers, Part II 4}}.
  Springer, \bibinfo{pages}{354--363}.
\newblock


\bibitem[\protect\citeauthoryear{Everingham, Van~Gool, Williams, Winn, and
  Zisserman}{Everingham et~al\mbox{.}}{2010}]%
        {everingham2010pascal}
\bibfield{author}{\bibinfo{person}{Mark Everingham}, \bibinfo{person}{Luc
  Van~Gool}, \bibinfo{person}{Christopher~KI Williams}, \bibinfo{person}{John
  Winn}, {and} \bibinfo{person}{Andrew Zisserman}.}
  \bibinfo{year}{2010}\natexlab{}.
\newblock \showarticletitle{The pascal visual object classes (voc) challenge}.
\newblock \bibinfo{journal}{\emph{International journal of computer vision}}
  \bibinfo{volume}{88}, \bibinfo{number}{2} (\bibinfo{year}{2010}),
  \bibinfo{pages}{303--338}.
\newblock


\bibitem[\protect\citeauthoryear{Fan and Ling}{Fan and Ling}{2019}]%
        {fan2019siamese}
\bibfield{author}{\bibinfo{person}{Heng Fan} {and} \bibinfo{person}{Haibin
  Ling}.} \bibinfo{year}{2019}\natexlab{}.
\newblock \showarticletitle{Siamese cascaded region proposal networks for
  real-time visual tracking}. In \bibinfo{booktitle}{\emph{Proceedings of the
  IEEE Conference on Computer Vision and Pattern Recognition}}.
  \bibinfo{pages}{7952--7961}.
\newblock


\bibitem[\protect\citeauthoryear{Farneb{\"a}ck}{Farneb{\"a}ck}{2003}]%
        {farneback2003two}
\bibfield{author}{\bibinfo{person}{Gunnar Farneb{\"a}ck}.}
  \bibinfo{year}{2003}\natexlab{}.
\newblock \showarticletitle{Two-frame motion estimation based on polynomial
  expansion}. In \bibinfo{booktitle}{\emph{Scandinavian conference on Image
  analysis}}. Springer, \bibinfo{pages}{363--370}.
\newblock


\bibitem[\protect\citeauthoryear{Girshick, Donahue, Darrell, and
  Malik}{Girshick et~al\mbox{.}}{2014}]%
        {girshick2014rich}
\bibfield{author}{\bibinfo{person}{Ross Girshick}, \bibinfo{person}{Jeff
  Donahue}, \bibinfo{person}{Trevor Darrell}, {and} \bibinfo{person}{Jitendra
  Malik}.} \bibinfo{year}{2014}\natexlab{}.
\newblock \showarticletitle{Rich feature hierarchies for accurate object
  detection and semantic segmentation}. In
  \bibinfo{booktitle}{\emph{Proceedings of the IEEE conference on computer
  vision and pattern recognition}}. \bibinfo{pages}{580--587}.
\newblock


\bibitem[\protect\citeauthoryear{Gong, Xiao, Tan, Sui, Xu, Duan, and Li}{Gong
  et~al\mbox{.}}{2019}]%
        {gong2019context}
\bibfield{author}{\bibinfo{person}{Yiping Gong}, \bibinfo{person}{Zhifeng
  Xiao}, \bibinfo{person}{Xiaowei Tan}, \bibinfo{person}{Haigang Sui},
  \bibinfo{person}{Chuan Xu}, \bibinfo{person}{Haiwang Duan}, {and}
  \bibinfo{person}{Deren Li}.} \bibinfo{year}{2019}\natexlab{}.
\newblock \showarticletitle{Context-Aware Convolutional Neural Network for
  Object Detection in VHR Remote Sensing Imagery}.
\newblock \bibinfo{journal}{\emph{IEEE Transactions on Geoscience and Remote
  Sensing}} \bibinfo{volume}{58}, \bibinfo{number}{1} (\bibinfo{year}{2019}),
  \bibinfo{pages}{34--44}.
\newblock


\bibitem[\protect\citeauthoryear{He, Gkioxari, Doll{\'a}r, and Girshick}{He
  et~al\mbox{.}}{2017}]%
        {he2017mask}
\bibfield{author}{\bibinfo{person}{Kaiming He}, \bibinfo{person}{Georgia
  Gkioxari}, \bibinfo{person}{Piotr Doll{\'a}r}, {and} \bibinfo{person}{Ross
  Girshick}.} \bibinfo{year}{2017}\natexlab{}.
\newblock \showarticletitle{Mask r-cnn}. In
  \bibinfo{booktitle}{\emph{Proceedings of the IEEE international conference on
  computer vision}}. \bibinfo{pages}{2961--2969}.
\newblock


\bibitem[\protect\citeauthoryear{He, Zhang, Ren, and Sun}{He
  et~al\mbox{.}}{2015}]%
        {he2015spatial}
\bibfield{author}{\bibinfo{person}{Kaiming He}, \bibinfo{person}{Xiangyu
  Zhang}, \bibinfo{person}{Shaoqing Ren}, {and} \bibinfo{person}{Jian Sun}.}
  \bibinfo{year}{2015}\natexlab{}.
\newblock \showarticletitle{Spatial pyramid pooling in deep convolutional
  networks for visual recognition}.
\newblock \bibinfo{journal}{\emph{IEEE transactions on pattern analysis and
  machine intelligence}} \bibinfo{volume}{37}, \bibinfo{number}{9}
  (\bibinfo{year}{2015}), \bibinfo{pages}{1904--1916}.
\newblock


\bibitem[\protect\citeauthoryear{He, Zhang, Ren, and Sun}{He
  et~al\mbox{.}}{2016}]%
        {he2016deep}
\bibfield{author}{\bibinfo{person}{Kaiming He}, \bibinfo{person}{Xiangyu
  Zhang}, \bibinfo{person}{Shaoqing Ren}, {and} \bibinfo{person}{Jian Sun}.}
  \bibinfo{year}{2016}\natexlab{}.
\newblock \showarticletitle{Deep residual learning for image recognition}. In
  \bibinfo{booktitle}{\emph{Proceedings of the IEEE conference on computer
  vision and pattern recognition}}. \bibinfo{pages}{770--778}.
\newblock


\bibitem[\protect\citeauthoryear{Hosang, Benenson, and Schiele}{Hosang
  et~al\mbox{.}}{2017}]%
        {hosang2017learning}
\bibfield{author}{\bibinfo{person}{Jan Hosang}, \bibinfo{person}{Rodrigo
  Benenson}, {and} \bibinfo{person}{Bernt Schiele}.}
  \bibinfo{year}{2017}\natexlab{}.
\newblock \showarticletitle{Learning non-maximum suppression}. In
  \bibinfo{booktitle}{\emph{Proceedings of the IEEE conference on computer
  vision and pattern recognition}}. \bibinfo{pages}{4507--4515}.
\newblock


\bibitem[\protect\citeauthoryear{Hsieh, Lin, and Hsu}{Hsieh
  et~al\mbox{.}}{2017}]%
        {hsieh2017drone}
\bibfield{author}{\bibinfo{person}{Meng-Ru Hsieh}, \bibinfo{person}{Yen-Liang
  Lin}, {and} \bibinfo{person}{Winston~H Hsu}.}
  \bibinfo{year}{2017}\natexlab{}.
\newblock \showarticletitle{Drone-based object counting by spatially
  regularized regional proposal network}. In
  \bibinfo{booktitle}{\emph{Proceedings of the IEEE International Conference on
  Computer Vision}}. \bibinfo{pages}{4145--4153}.
\newblock


\bibitem[\protect\citeauthoryear{Jain, Xiong, and Grauman}{Jain
  et~al\mbox{.}}{2017}]%
        {jain2017fusionseg}
\bibfield{author}{\bibinfo{person}{Suyog~Dutt Jain}, \bibinfo{person}{Bo
  Xiong}, {and} \bibinfo{person}{Kristen Grauman}.}
  \bibinfo{year}{2017}\natexlab{}.
\newblock \showarticletitle{Fusionseg: Learning to combine motion and
  appearance for fully automatic segmentation of generic objects in videos}. In
  \bibinfo{booktitle}{\emph{2017 IEEE conference on computer vision and pattern
  recognition (CVPR)}}. IEEE, \bibinfo{pages}{2117--2126}.
\newblock


\bibitem[\protect\citeauthoryear{Koh and Kim}{Koh and Kim}{2017}]%
        {koh2017primary}
\bibfield{author}{\bibinfo{person}{Yeong~Jun Koh} {and}
  \bibinfo{person}{Chang-Su Kim}.} \bibinfo{year}{2017}\natexlab{}.
\newblock \showarticletitle{Primary object segmentation in videos based on
  region augmentation and reduction}. In \bibinfo{booktitle}{\emph{2017 IEEE
  Conference on Computer Vision and Pattern Recognition (CVPR)}}. IEEE,
  \bibinfo{pages}{7417--7425}.
\newblock


\bibitem[\protect\citeauthoryear{Kong, Sun, Yao, Liu, Lu, and Chen}{Kong
  et~al\mbox{.}}{2017}]%
        {kong2017ron}
\bibfield{author}{\bibinfo{person}{Tao Kong}, \bibinfo{person}{Fuchun Sun},
  \bibinfo{person}{Anbang Yao}, \bibinfo{person}{Huaping Liu},
  \bibinfo{person}{Ming Lu}, {and} \bibinfo{person}{Yurong Chen}.}
  \bibinfo{year}{2017}\natexlab{}.
\newblock \showarticletitle{Ron: Reverse connection with objectness prior
  networks for object detection}. In \bibinfo{booktitle}{\emph{Proceedings of
  the IEEE conference on computer vision and pattern recognition}}.
  \bibinfo{pages}{5936--5944}.
\newblock


\bibitem[\protect\citeauthoryear{Law and Deng}{Law and Deng}{2018}]%
        {law2018cornernet}
\bibfield{author}{\bibinfo{person}{Hei Law} {and} \bibinfo{person}{Jia Deng}.}
  \bibinfo{year}{2018}\natexlab{}.
\newblock \showarticletitle{Cornernet: Detecting objects as paired keypoints}.
  In \bibinfo{booktitle}{\emph{Proceedings of the European Conference on
  Computer Vision (ECCV)}}. \bibinfo{pages}{734--750}.
\newblock


\bibitem[\protect\citeauthoryear{Lezki, Ahu~Ozturk, Akif~Akpinar, Kerim~Yucel,
  Berker~Logoglu, Erdem, and Erdem}{Lezki et~al\mbox{.}}{2018}]%
        {lezki2018joint}
\bibfield{author}{\bibinfo{person}{Hazal Lezki}, \bibinfo{person}{I
  Ahu~Ozturk}, \bibinfo{person}{M Akif~Akpinar}, \bibinfo{person}{M
  Kerim~Yucel}, \bibinfo{person}{K Berker~Logoglu}, \bibinfo{person}{Aykut
  Erdem}, {and} \bibinfo{person}{Erkut Erdem}.}
  \bibinfo{year}{2018}\natexlab{}.
\newblock \showarticletitle{Joint exploitation of features and optical flow for
  real-time moving object detection on drones}. In
  \bibinfo{booktitle}{\emph{Proceedings of the European Conference on Computer
  Vision (ECCV)}}. \bibinfo{pages}{0--0}.
\newblock


\bibitem[\protect\citeauthoryear{Li, Ye, Chung, Kolsch, Wachs, and Bouman}{Li
  et~al\mbox{.}}{2016}]%
        {li2016multi}
\bibfield{author}{\bibinfo{person}{Jing Li}, \bibinfo{person}{Dong~Hye Ye},
  \bibinfo{person}{Timothy Chung}, \bibinfo{person}{Mathias Kolsch},
  \bibinfo{person}{Juan Wachs}, {and} \bibinfo{person}{Charles Bouman}.}
  \bibinfo{year}{2016}\natexlab{}.
\newblock \showarticletitle{Multi-target detection and tracking from a single
  camera in Unmanned Aerial Vehicles (UAVs)}. In \bibinfo{booktitle}{\emph{2016
  IEEE/RSJ International Conference on Intelligent Robots and Systems (IROS)}}.
  IEEE, \bibinfo{pages}{4992--4997}.
\newblock


\bibitem[\protect\citeauthoryear{Liang, Zhang, Zhuo, Li, and Tian}{Liang
  et~al\mbox{.}}{2019}]%
        {liang2019small}
\bibfield{author}{\bibinfo{person}{Xi Liang}, \bibinfo{person}{Jing Zhang},
  \bibinfo{person}{Li Zhuo}, \bibinfo{person}{Yuzhao Li}, {and}
  \bibinfo{person}{Qi Tian}.} \bibinfo{year}{2019}\natexlab{}.
\newblock \showarticletitle{Small Object Detection in Unmanned Aerial Vehicle
  Images Using Feature Fusion and Scaling-Based Single Shot Detector with
  Spatial Context Analysis}.
\newblock \bibinfo{journal}{\emph{IEEE Transactions on Circuits and Systems for
  Video Technology}} (\bibinfo{year}{2019}).
\newblock


\bibitem[\protect\citeauthoryear{Lin, Chen, Santoso, Lin, and Lai}{Lin
  et~al\mbox{.}}{2019}]%
        {lin2019real}
\bibfield{author}{\bibinfo{person}{Che-Tsung Lin}, \bibinfo{person}{Shu-Ping
  Chen}, \bibinfo{person}{Patrisia~Sherryl Santoso}, \bibinfo{person}{Hung-Jin
  Lin}, {and} \bibinfo{person}{Shang-Hong Lai}.}
  \bibinfo{year}{2019}\natexlab{}.
\newblock \showarticletitle{Real-time Single-Stage Vehicle Detector Optimized
  by Multi-Stage Image-based Online Hard Example Mining}.
\newblock \bibinfo{journal}{\emph{IEEE Transactions on Vehicular Technology}}
  (\bibinfo{year}{2019}).
\newblock


\bibitem[\protect\citeauthoryear{{Lin}, {Goyal}, {Girshick}, {He}, and
  {Dollár}}{{Lin} et~al\mbox{.}}{2020}]%
        {linfocalpami}
\bibfield{author}{\bibinfo{person}{T. {Lin}}, \bibinfo{person}{P. {Goyal}},
  \bibinfo{person}{R. {Girshick}}, \bibinfo{person}{K. {He}}, {and}
  \bibinfo{person}{P. {Dollár}}.} \bibinfo{year}{2020}\natexlab{}.
\newblock \showarticletitle{Focal Loss for Dense Object Detection}.
\newblock \bibinfo{journal}{\emph{IEEE Transactions on Pattern Analysis and
  Machine Intelligence}} \bibinfo{volume}{42}, \bibinfo{number}{2}
  (\bibinfo{date}{Feb} \bibinfo{year}{2020}), \bibinfo{pages}{318--327}.
\newblock
\showISSN{1939-3539}
\urldef\tempurl%
\url{https://doi.org/10.1109/TPAMI.2018.2858826}
\showDOI{\tempurl}


\bibitem[\protect\citeauthoryear{Lin, Doll{\'a}r, Girshick, He, Hariharan, and
  Belongie}{Lin et~al\mbox{.}}{2017}]%
        {lin2017feature}
\bibfield{author}{\bibinfo{person}{Tsung-Yi Lin}, \bibinfo{person}{Piotr
  Doll{\'a}r}, \bibinfo{person}{Ross Girshick}, \bibinfo{person}{Kaiming He},
  \bibinfo{person}{Bharath Hariharan}, {and} \bibinfo{person}{Serge Belongie}.}
  \bibinfo{year}{2017}\natexlab{}.
\newblock \showarticletitle{Feature pyramid networks for object detection}. In
  \bibinfo{booktitle}{\emph{Proceedings of the IEEE conference on computer
  vision and pattern recognition}}. \bibinfo{pages}{2117--2125}.
\newblock


\bibitem[\protect\citeauthoryear{Lin, Maire, Belongie, Hays, Perona, Ramanan,
  Doll{\'a}r, and Zitnick}{Lin et~al\mbox{.}}{2014}]%
        {lin2014microsoft}
\bibfield{author}{\bibinfo{person}{Tsung-Yi Lin}, \bibinfo{person}{Michael
  Maire}, \bibinfo{person}{Serge Belongie}, \bibinfo{person}{James Hays},
  \bibinfo{person}{Pietro Perona}, \bibinfo{person}{Deva Ramanan},
  \bibinfo{person}{Piotr Doll{\'a}r}, {and} \bibinfo{person}{C~Lawrence
  Zitnick}.} \bibinfo{year}{2014}\natexlab{}.
\newblock \showarticletitle{Microsoft coco: Common objects in context}. In
  \bibinfo{booktitle}{\emph{European conference on computer vision}}. Springer,
  \bibinfo{pages}{740--755}.
\newblock


\bibitem[\protect\citeauthoryear{Liu and Mattyus}{Liu and Mattyus}{2015}]%
        {liu2015fast}
\bibfield{author}{\bibinfo{person}{Kang Liu} {and} \bibinfo{person}{Gellert
  Mattyus}.} \bibinfo{year}{2015}\natexlab{}.
\newblock \showarticletitle{Fast multiclass vehicle detection on aerial
  images}.
\newblock \bibinfo{journal}{\emph{IEEE Geoscience and Remote Sensing Letters}}
  \bibinfo{volume}{12}, \bibinfo{number}{9} (\bibinfo{year}{2015}),
  \bibinfo{pages}{1938--1942}.
\newblock


\bibitem[\protect\citeauthoryear{Liu, Anguelov, Erhan, Szegedy, Reed, Fu, and
  Berg}{Liu et~al\mbox{.}}{2016}]%
        {liu2016ssd}
\bibfield{author}{\bibinfo{person}{Wei Liu}, \bibinfo{person}{Dragomir
  Anguelov}, \bibinfo{person}{Dumitru Erhan}, \bibinfo{person}{Christian
  Szegedy}, \bibinfo{person}{Scott Reed}, \bibinfo{person}{Cheng-Yang Fu},
  {and} \bibinfo{person}{Alexander~C Berg}.} \bibinfo{year}{2016}\natexlab{}.
\newblock \showarticletitle{Ssd: Single shot multibox detector}. In
  \bibinfo{booktitle}{\emph{European conference on computer vision}}. Springer,
  \bibinfo{pages}{21--37}.
\newblock


\bibitem[\protect\citeauthoryear{Liu, Ma, Wang, et~al\mbox{.}}{Liu
  et~al\mbox{.}}{2018}]%
        {liu2018detection}
\bibfield{author}{\bibinfo{person}{Wenchao Liu}, \bibinfo{person}{Long Ma},
  \bibinfo{person}{Jue Wang}, {et~al\mbox{.}}} \bibinfo{year}{2018}\natexlab{}.
\newblock \showarticletitle{Detection of multiclass objects in optical remote
  sensing images}.
\newblock \bibinfo{journal}{\emph{IEEE Geoscience and Remote Sensing Letters}}
  \bibinfo{volume}{16}, \bibinfo{number}{5} (\bibinfo{year}{2018}),
  \bibinfo{pages}{791--795}.
\newblock


\bibitem[\protect\citeauthoryear{Mandal, Chaudhary, Vipparthi, Murala, Gonde,
  and Nagar}{Mandal et~al\mbox{.}}{2018a}]%
        {mandal2018antic}
\bibfield{author}{\bibinfo{person}{Murari Mandal}, \bibinfo{person}{Mallika
  Chaudhary}, \bibinfo{person}{Santosh~Kumar Vipparthi},
  \bibinfo{person}{Subrahmanyam Murala}, \bibinfo{person}{Anil~Balaji Gonde},
  {and} \bibinfo{person}{Shyam~Krishna Nagar}.}
  \bibinfo{year}{2018}\natexlab{a}.
\newblock \showarticletitle{ANTIC: ANTithetic isomeric cluster patterns for
  medical image retrieval and change detection}.
\newblock \bibinfo{journal}{\emph{IET Computer Vision}} \bibinfo{volume}{13},
  \bibinfo{number}{1} (\bibinfo{year}{2018}), \bibinfo{pages}{31--43}.
\newblock


\bibitem[\protect\citeauthoryear{Mandal, Dhar, Mishra, and Vipparthi}{Mandal
  et~al\mbox{.}}{2019a}]%
        {mandal20193dfr}
\bibfield{author}{\bibinfo{person}{Murari Mandal}, \bibinfo{person}{Vansh
  Dhar}, \bibinfo{person}{Abhishek Mishra}, {and}
  \bibinfo{person}{Santosh~Kumar Vipparthi}.} \bibinfo{year}{2019}\natexlab{a}.
\newblock \showarticletitle{3DFR: A Swift 3D Feature Reductionist Framework for
  Scene Independent Change Detection}.
\newblock \bibinfo{journal}{\emph{IEEE Signal Processing Letters}}
  \bibinfo{volume}{26}, \bibinfo{number}{12} (\bibinfo{year}{2019}),
  \bibinfo{pages}{1882--1886}.
\newblock


\bibitem[\protect\citeauthoryear{Mandal, Kumar, Saran, and vipparthi}{Mandal
  et~al\mbox{.}}{2020}]%
        {Mandal_2020_WACV}
\bibfield{author}{\bibinfo{person}{Murari Mandal}, \bibinfo{person}{Lav~Kush
  Kumar}, \bibinfo{person}{Mahipal~Singh Saran}, {and}
  \bibinfo{person}{Santosh~Kumar vipparthi}.} \bibinfo{year}{2020}\natexlab{}.
\newblock \showarticletitle{MotionRec: A Unified Deep Framework for Moving
  Object Recognition}. In \bibinfo{booktitle}{\emph{The IEEE Winter Conference
  on Applications of Computer Vision (WACV)}}.
\newblock


\bibitem[\protect\citeauthoryear{Mandal, Saxena, Vipparthi, and Murala}{Mandal
  et~al\mbox{.}}{2018b}]%
        {mandal2018candid}
\bibfield{author}{\bibinfo{person}{Murari Mandal}, \bibinfo{person}{Prafulla
  Saxena}, \bibinfo{person}{Santosh~Kumar Vipparthi}, {and}
  \bibinfo{person}{Subrahmanyam Murala}.} \bibinfo{year}{2018}\natexlab{b}.
\newblock \showarticletitle{CANDID: Robust change dynamics and deterministic
  update policy for dynamic background subtraction}. In
  \bibinfo{booktitle}{\emph{2018 24th International Conference on Pattern
  Recognition (ICPR)}}. IEEE, \bibinfo{pages}{2468--2473}.
\newblock


\bibitem[\protect\citeauthoryear{Mandal, Shah, Meena, Devi, and
  Vipparthi}{Mandal et~al\mbox{.}}{2019c}]%
        {mandal2019avdnet}
\bibfield{author}{\bibinfo{person}{Murari Mandal}, \bibinfo{person}{Manal
  Shah}, \bibinfo{person}{Prashant Meena}, \bibinfo{person}{Sanhita Devi},
  {and} \bibinfo{person}{Santosh~Kumar Vipparthi}.}
  \bibinfo{year}{2019}\natexlab{c}.
\newblock \showarticletitle{AVDNet: A Small-Sized Vehicle Detection Network for
  Aerial Visual Data}.
\newblock \bibinfo{journal}{\emph{IEEE Geoscience and Remote Sensing Letters}}
  (\bibinfo{year}{2019}).
\newblock


\bibitem[\protect\citeauthoryear{Mandal, Shah, Meena, and Vipparthi}{Mandal
  et~al\mbox{.}}{2019b}]%
        {mandal2019sssdet}
\bibfield{author}{\bibinfo{person}{Murari Mandal}, \bibinfo{person}{Manal
  Shah}, \bibinfo{person}{Prashant Meena}, {and} \bibinfo{person}{Santosh~Kumar
  Vipparthi}.} \bibinfo{year}{2019}\natexlab{b}.
\newblock \showarticletitle{SSSDET: Simple Short and Shallow Network for
  Resource Efficient Vehicle Detection in Aerial Scenes}. In
  \bibinfo{booktitle}{\emph{2019 IEEE International Conference on Image
  Processing (ICIP)}}. IEEE, \bibinfo{pages}{3098--3102}.
\newblock


\bibitem[\protect\citeauthoryear{Maninis, Caelles, Chen, Pont-Tuset,
  Leal-Taix{\'e}, Cremers, and Van~Gool}{Maninis et~al\mbox{.}}{2018}]%
        {maninis2018video}
\bibfield{author}{\bibinfo{person}{K-K Maninis}, \bibinfo{person}{Sergi
  Caelles}, \bibinfo{person}{Yuhua Chen}, \bibinfo{person}{Jordi Pont-Tuset},
  \bibinfo{person}{Laura Leal-Taix{\'e}}, \bibinfo{person}{Daniel Cremers},
  {and} \bibinfo{person}{Luc Van~Gool}.} \bibinfo{year}{2018}\natexlab{}.
\newblock \showarticletitle{Video object segmentation without temporal
  information}.
\newblock \bibinfo{journal}{\emph{IEEE transactions on pattern analysis and
  machine intelligence}} \bibinfo{volume}{41}, \bibinfo{number}{6}
  (\bibinfo{year}{2018}), \bibinfo{pages}{1515--1530}.
\newblock


\bibitem[\protect\citeauthoryear{Marotirao~Biradar, Gupta, Mandal, and
  Kumar~Vipparthi}{Marotirao~Biradar et~al\mbox{.}}{2019}]%
        {marotirao2019challenges}
\bibfield{author}{\bibinfo{person}{Kuldeep Marotirao~Biradar},
  \bibinfo{person}{Ayushi Gupta}, \bibinfo{person}{Murari Mandal}, {and}
  \bibinfo{person}{Santosh Kumar~Vipparthi}.} \bibinfo{year}{2019}\natexlab{}.
\newblock \showarticletitle{Challenges in Time-Stamp Aware Anomaly Detection in
  Traffic Videos}. In \bibinfo{booktitle}{\emph{Proceedings of the IEEE
  Conference on Computer Vision and Pattern Recognition Workshops}}.
  \bibinfo{pages}{13--20}.
\newblock


\bibitem[\protect\citeauthoryear{Mehta, Sinha, Narang, and Mandal}{Mehta
  et~al\mbox{.}}{2020}]%
        {mehta2020hidegan}
\bibfield{author}{\bibinfo{person}{Aditya Mehta}, \bibinfo{person}{Harsh
  Sinha}, \bibinfo{person}{Pratik Narang}, {and} \bibinfo{person}{Murari
  Mandal}.} \bibinfo{year}{2020}\natexlab{}.
\newblock \showarticletitle{HIDeGan: A Hyperspectral-Guided Image Dehazing
  GAN}. In \bibinfo{booktitle}{\emph{Proceedings of the IEEE/CVF Conference on
  Computer Vision and Pattern Recognition Workshops}}.
  \bibinfo{pages}{212--213}.
\newblock


\bibitem[\protect\citeauthoryear{Mueller, Smith, and Ghanem}{Mueller
  et~al\mbox{.}}{2016}]%
        {mueller2016benchmark}
\bibfield{author}{\bibinfo{person}{Matthias Mueller}, \bibinfo{person}{Neil
  Smith}, {and} \bibinfo{person}{Bernard Ghanem}.}
  \bibinfo{year}{2016}\natexlab{}.
\newblock \showarticletitle{A benchmark and simulator for uav tracking}. In
  \bibinfo{booktitle}{\emph{European conference on computer vision}}. Springer,
  \bibinfo{pages}{445--461}.
\newblock


\bibitem[\protect\citeauthoryear{Ochs, Malik, and Brox}{Ochs
  et~al\mbox{.}}{2013}]%
        {ochs2013segmentation}
\bibfield{author}{\bibinfo{person}{Peter Ochs}, \bibinfo{person}{Jitendra
  Malik}, {and} \bibinfo{person}{Thomas Brox}.}
  \bibinfo{year}{2013}\natexlab{}.
\newblock \showarticletitle{Segmentation of moving objects by long term video
  analysis}.
\newblock \bibinfo{journal}{\emph{IEEE transactions on pattern analysis and
  machine intelligence}} \bibinfo{volume}{36}, \bibinfo{number}{6}
  (\bibinfo{year}{2013}), \bibinfo{pages}{1187--1200}.
\newblock


\bibitem[\protect\citeauthoryear{Pan, Tong, Zhao, Zhao, Su, and Zhuang}{Pan
  et~al\mbox{.}}{2019}]%
        {pan2019multi}
\bibfield{author}{\bibinfo{person}{Siyang Pan}, \bibinfo{person}{Zhihang Tong},
  \bibinfo{person}{Yanyun Zhao}, \bibinfo{person}{Zhicheng Zhao},
  \bibinfo{person}{Fei Su}, {and} \bibinfo{person}{Bojin Zhuang}.}
  \bibinfo{year}{2019}\natexlab{}.
\newblock \showarticletitle{Multi-Object Tracking Hierarchically in Visual Data
  Taken From Drones}. In \bibinfo{booktitle}{\emph{Proceedings of the IEEE
  International Conference on Computer Vision Workshops}}.
  \bibinfo{pages}{0--0}.
\newblock


\bibitem[\protect\citeauthoryear{Papazoglou and Ferrari}{Papazoglou and
  Ferrari}{2013}]%
        {papazoglou2013fast}
\bibfield{author}{\bibinfo{person}{Anestis Papazoglou} {and}
  \bibinfo{person}{Vittorio Ferrari}.} \bibinfo{year}{2013}\natexlab{}.
\newblock \showarticletitle{Fast object segmentation in unconstrained video}.
  In \bibinfo{booktitle}{\emph{Proceedings of the IEEE international conference
  on computer vision}}. \bibinfo{pages}{1777--1784}.
\newblock


\bibitem[\protect\citeauthoryear{Pont-Tuset, Perazzi, Caelles, Arbel{\'a}ez,
  Sorkine-Hornung, and Van~Gool}{Pont-Tuset et~al\mbox{.}}{2017}]%
        {pont20172017}
\bibfield{author}{\bibinfo{person}{Jordi Pont-Tuset}, \bibinfo{person}{Federico
  Perazzi}, \bibinfo{person}{Sergi Caelles}, \bibinfo{person}{Pablo
  Arbel{\'a}ez}, \bibinfo{person}{Alex Sorkine-Hornung}, {and}
  \bibinfo{person}{Luc Van~Gool}.} \bibinfo{year}{2017}\natexlab{}.
\newblock \showarticletitle{The 2017 davis challenge on video object
  segmentation}.
\newblock \bibinfo{journal}{\emph{arXiv preprint arXiv:1704.00675}}
  (\bibinfo{year}{2017}).
\newblock


\bibitem[\protect\citeauthoryear{Razakarivony and Jurie}{Razakarivony and
  Jurie}{2016}]%
        {razakarivony2016vehicle}
\bibfield{author}{\bibinfo{person}{Sebastien Razakarivony} {and}
  \bibinfo{person}{Frederic Jurie}.} \bibinfo{year}{2016}\natexlab{}.
\newblock \showarticletitle{Vehicle detection in aerial imagery: A small target
  detection benchmark}.
\newblock \bibinfo{journal}{\emph{Journal of Visual Communication and Image
  Representation}}  \bibinfo{volume}{34} (\bibinfo{year}{2016}),
  \bibinfo{pages}{187--203}.
\newblock


\bibitem[\protect\citeauthoryear{{Ren}, {He}, {Girshick}, and {Sun}}{{Ren}
  et~al\mbox{.}}{2017}]%
        {7485869}
\bibfield{author}{\bibinfo{person}{S. {Ren}}, \bibinfo{person}{K. {He}},
  \bibinfo{person}{R. {Girshick}}, {and} \bibinfo{person}{J. {Sun}}.}
  \bibinfo{year}{2017}\natexlab{}.
\newblock \showarticletitle{Faster R-CNN: Towards Real-Time Object Detection
  with Region Proposal Networks}.
\newblock \bibinfo{journal}{\emph{IEEE Transactions on Pattern Analysis and
  Machine Intelligence}} \bibinfo{volume}{39}, \bibinfo{number}{6}
  (\bibinfo{date}{June} \bibinfo{year}{2017}), \bibinfo{pages}{1137--1149}.
\newblock
\showISSN{1939-3539}
\urldef\tempurl%
\url{https://doi.org/10.1109/TPAMI.2016.2577031}
\showDOI{\tempurl}


\bibitem[\protect\citeauthoryear{Robicquet, Sadeghian, Alahi, and
  Savarese}{Robicquet et~al\mbox{.}}{2016}]%
        {robicquet2016learning}
\bibfield{author}{\bibinfo{person}{Alexandre Robicquet}, \bibinfo{person}{Amir
  Sadeghian}, \bibinfo{person}{Alexandre Alahi}, {and} \bibinfo{person}{Silvio
  Savarese}.} \bibinfo{year}{2016}\natexlab{}.
\newblock \showarticletitle{Learning social etiquette: Human trajectory
  understanding in crowded scenes}. In \bibinfo{booktitle}{\emph{European
  conference on computer vision}}. Springer, \bibinfo{pages}{549--565}.
\newblock


\bibitem[\protect\citeauthoryear{Sandler, Howard, Zhu, Zhmoginov, and
  Chen}{Sandler et~al\mbox{.}}{2018}]%
        {sandler2018mobilenetv2}
\bibfield{author}{\bibinfo{person}{Mark Sandler}, \bibinfo{person}{Andrew
  Howard}, \bibinfo{person}{Menglong Zhu}, \bibinfo{person}{Andrey Zhmoginov},
  {and} \bibinfo{person}{Liang-Chieh Chen}.} \bibinfo{year}{2018}\natexlab{}.
\newblock \showarticletitle{Mobilenetv2: Inverted residuals and linear
  bottlenecks}. In \bibinfo{booktitle}{\emph{Proceedings of the IEEE conference
  on computer vision and pattern recognition}}. \bibinfo{pages}{4510--4520}.
\newblock


\bibitem[\protect\citeauthoryear{Sermanet, Eigen, Zhang, Mathieu, Fergus, and
  LeCun}{Sermanet et~al\mbox{.}}{2013}]%
        {sermanet2013overfeat}
\bibfield{author}{\bibinfo{person}{Pierre Sermanet}, \bibinfo{person}{David
  Eigen}, \bibinfo{person}{Xiang Zhang}, \bibinfo{person}{Micha{\"e}l Mathieu},
  \bibinfo{person}{Rob Fergus}, {and} \bibinfo{person}{Yann LeCun}.}
  \bibinfo{year}{2013}\natexlab{}.
\newblock \showarticletitle{Overfeat: Integrated recognition, localization and
  detection using convolutional networks}.
\newblock \bibinfo{journal}{\emph{arXiv preprint arXiv:1312.6229}}
  (\bibinfo{year}{2013}).
\newblock


\bibitem[\protect\citeauthoryear{Shrivastava, Gupta, and Girshick}{Shrivastava
  et~al\mbox{.}}{2016}]%
        {shrivastava2016training}
\bibfield{author}{\bibinfo{person}{Abhinav Shrivastava},
  \bibinfo{person}{Abhinav Gupta}, {and} \bibinfo{person}{Ross Girshick}.}
  \bibinfo{year}{2016}\natexlab{}.
\newblock \showarticletitle{Training region-based object detectors with online
  hard example mining}. In \bibinfo{booktitle}{\emph{Proceedings of the IEEE
  conference on computer vision and pattern recognition}}.
  \bibinfo{pages}{761--769}.
\newblock


\bibitem[\protect\citeauthoryear{Singh, Najibi, and Davis}{Singh
  et~al\mbox{.}}{2018}]%
        {singh2018sniper}
\bibfield{author}{\bibinfo{person}{Bharat Singh}, \bibinfo{person}{Mahyar
  Najibi}, {and} \bibinfo{person}{Larry~S Davis}.}
  \bibinfo{year}{2018}\natexlab{}.
\newblock \showarticletitle{SNIPER: Efficient multi-scale training}. In
  \bibinfo{booktitle}{\emph{Advances in neural information processing
  systems}}. \bibinfo{pages}{9310--9320}.
\newblock


\bibitem[\protect\citeauthoryear{Song, Wang, Zhao, Shen, and Lam}{Song
  et~al\mbox{.}}{2018}]%
        {song2018pyramid}
\bibfield{author}{\bibinfo{person}{Hongmei Song}, \bibinfo{person}{Wenguan
  Wang}, \bibinfo{person}{Sanyuan Zhao}, \bibinfo{person}{Jianbing Shen}, {and}
  \bibinfo{person}{Kin-Man Lam}.} \bibinfo{year}{2018}\natexlab{}.
\newblock \showarticletitle{Pyramid dilated deeper convlstm for video salient
  object detection}. In \bibinfo{booktitle}{\emph{Proceedings of the European
  Conference on Computer Vision (ECCV)}}. \bibinfo{pages}{715--731}.
\newblock


\bibitem[\protect\citeauthoryear{Tan, Chen, Pang, Vasudevan, Sandler, Howard,
  and Le}{Tan et~al\mbox{.}}{2019}]%
        {tan2019mnasnet}
\bibfield{author}{\bibinfo{person}{Mingxing Tan}, \bibinfo{person}{Bo Chen},
  \bibinfo{person}{Ruoming Pang}, \bibinfo{person}{Vijay Vasudevan},
  \bibinfo{person}{Mark Sandler}, \bibinfo{person}{Andrew Howard}, {and}
  \bibinfo{person}{Quoc~V Le}.} \bibinfo{year}{2019}\natexlab{}.
\newblock \showarticletitle{Mnasnet: Platform-aware neural architecture search
  for mobile}. In \bibinfo{booktitle}{\emph{Proceedings of the IEEE Conference
  on Computer Vision and Pattern Recognition}}. \bibinfo{pages}{2820--2828}.
\newblock


\bibitem[\protect\citeauthoryear{Valmadre, Bertinetto, Henriques, Vedaldi, and
  Torr}{Valmadre et~al\mbox{.}}{2017}]%
        {valmadre2017end}
\bibfield{author}{\bibinfo{person}{Jack Valmadre}, \bibinfo{person}{Luca
  Bertinetto}, \bibinfo{person}{Joao Henriques}, \bibinfo{person}{Andrea
  Vedaldi}, {and} \bibinfo{person}{Philip~HS Torr}.}
  \bibinfo{year}{2017}\natexlab{}.
\newblock \showarticletitle{End-to-end representation learning for correlation
  filter based tracking}. In \bibinfo{booktitle}{\emph{Proceedings of the IEEE
  Conference on Computer Vision and Pattern Recognition}}.
  \bibinfo{pages}{2805--2813}.
\newblock


\bibitem[\protect\citeauthoryear{Wang, Jodoin, Porikli, Konrad, Benezeth, and
  Ishwar}{Wang et~al\mbox{.}}{2014}]%
        {wang2014cdnet}
\bibfield{author}{\bibinfo{person}{Yi Wang}, \bibinfo{person}{Pierre-Marc
  Jodoin}, \bibinfo{person}{Fatih Porikli}, \bibinfo{person}{Janusz Konrad},
  \bibinfo{person}{Yannick Benezeth}, {and} \bibinfo{person}{Prakash Ishwar}.}
  \bibinfo{year}{2014}\natexlab{}.
\newblock \showarticletitle{CDnet 2014: An expanded change detection benchmark
  dataset}. In \bibinfo{booktitle}{\emph{Proceedings of the IEEE conference on
  computer vision and pattern recognition workshops}}.
  \bibinfo{pages}{387--394}.
\newblock


\bibitem[\protect\citeauthoryear{Wang, Shi, and Wu}{Wang et~al\mbox{.}}{2017}]%
        {wang2017robust}
\bibfield{author}{\bibinfo{person}{Yong Wang}, \bibinfo{person}{Wei Shi}, {and}
  \bibinfo{person}{Shandong Wu}.} \bibinfo{year}{2017}\natexlab{}.
\newblock \showarticletitle{Robust UAV-based tracking using hybrid
  classifiers}. In \bibinfo{booktitle}{\emph{Proceedings of the IEEE
  International Conference on Computer Vision Workshops}}.
  \bibinfo{pages}{2129--2137}.
\newblock


\bibitem[\protect\citeauthoryear{Waqas~Zamir, Arora, Gupta, Khan, Sun,
  Shahbaz~Khan, Zhu, Shao, Xia, and Bai}{Waqas~Zamir et~al\mbox{.}}{2019}]%
        {waqas2019isaid}
\bibfield{author}{\bibinfo{person}{Syed Waqas~Zamir}, \bibinfo{person}{Aditya
  Arora}, \bibinfo{person}{Akshita Gupta}, \bibinfo{person}{Salman Khan},
  \bibinfo{person}{Guolei Sun}, \bibinfo{person}{Fahad Shahbaz~Khan},
  \bibinfo{person}{Fan Zhu}, \bibinfo{person}{Ling Shao},
  \bibinfo{person}{Gui-Song Xia}, {and} \bibinfo{person}{Xiang Bai}.}
  \bibinfo{year}{2019}\natexlab{}.
\newblock \showarticletitle{iSAID: A Large-scale Dataset for Instance
  Segmentation in Aerial Images}. In \bibinfo{booktitle}{\emph{Proceedings of
  the IEEE Conference on Computer Vision and Pattern Recognition Workshops}}.
  \bibinfo{pages}{28--37}.
\newblock


\bibitem[\protect\citeauthoryear{Wu, Dai, Zhang, Wang, Sun, Wu, Tian, Vajda,
  Jia, and Keutzer}{Wu et~al\mbox{.}}{2019a}]%
        {wu2019fbnet}
\bibfield{author}{\bibinfo{person}{Bichen Wu}, \bibinfo{person}{Xiaoliang Dai},
  \bibinfo{person}{Peizhao Zhang}, \bibinfo{person}{Yanghan Wang},
  \bibinfo{person}{Fei Sun}, \bibinfo{person}{Yiming Wu},
  \bibinfo{person}{Yuandong Tian}, \bibinfo{person}{Peter Vajda},
  \bibinfo{person}{Yangqing Jia}, {and} \bibinfo{person}{Kurt Keutzer}.}
  \bibinfo{year}{2019}\natexlab{a}.
\newblock \showarticletitle{Fbnet: Hardware-aware efficient convnet design via
  differentiable neural architecture search}. In
  \bibinfo{booktitle}{\emph{Proceedings of the IEEE Conference on Computer
  Vision and Pattern Recognition}}. \bibinfo{pages}{10734--10742}.
\newblock


\bibitem[\protect\citeauthoryear{Wu, Hong, Tian, Chanussot, Li, and Tao}{Wu
  et~al\mbox{.}}{2019b}]%
        {wu2019orsim}
\bibfield{author}{\bibinfo{person}{Xin Wu}, \bibinfo{person}{Danfeng Hong},
  \bibinfo{person}{Jiaojiao Tian}, \bibinfo{person}{Jocelyn Chanussot},
  \bibinfo{person}{Wei Li}, {and} \bibinfo{person}{Ran Tao}.}
  \bibinfo{year}{2019}\natexlab{b}.
\newblock \showarticletitle{ORSIm Detector: A novel object detection framework
  in optical remote sensing imagery using spatial-frequency channel features}.
\newblock \bibinfo{journal}{\emph{IEEE Transactions on Geoscience and Remote
  Sensing}} \bibinfo{volume}{57}, \bibinfo{number}{7} (\bibinfo{year}{2019}),
  \bibinfo{pages}{5146--5158}.
\newblock


\bibitem[\protect\citeauthoryear{Wu, Suresh, Narayanan, Xu, Kwon, and Wang}{Wu
  et~al\mbox{.}}{2019c}]%
        {wu2019delving}
\bibfield{author}{\bibinfo{person}{Zhenyu Wu}, \bibinfo{person}{Karthik
  Suresh}, \bibinfo{person}{Priya Narayanan}, \bibinfo{person}{Hongyu Xu},
  \bibinfo{person}{Heesung Kwon}, {and} \bibinfo{person}{Zhangyang Wang}.}
  \bibinfo{year}{2019}\natexlab{c}.
\newblock \showarticletitle{Delving into robust object detection from unmanned
  aerial vehicles: A deep nuisance disentanglement approach}. In
  \bibinfo{booktitle}{\emph{Proceedings of the IEEE International Conference on
  Computer Vision}}. \bibinfo{pages}{1201--1210}.
\newblock


\bibitem[\protect\citeauthoryear{Xia, Bai, Ding, Zhu, Belongie, Luo, Datcu,
  Pelillo, and Zhang}{Xia et~al\mbox{.}}{2018}]%
        {xia2018dota}
\bibfield{author}{\bibinfo{person}{Gui-Song Xia}, \bibinfo{person}{Xiang Bai},
  \bibinfo{person}{Jian Ding}, \bibinfo{person}{Zhen Zhu},
  \bibinfo{person}{Serge Belongie}, \bibinfo{person}{Jiebo Luo},
  \bibinfo{person}{Mihai Datcu}, \bibinfo{person}{Marcello Pelillo}, {and}
  \bibinfo{person}{Liangpei Zhang}.} \bibinfo{year}{2018}\natexlab{}.
\newblock \showarticletitle{DOTA: A large-scale dataset for object detection in
  aerial images}. In \bibinfo{booktitle}{\emph{Proceedings of the IEEE
  Conference on Computer Vision and Pattern Recognition}}.
  \bibinfo{pages}{3974--3983}.
\newblock


\bibitem[\protect\citeauthoryear{Xu, Yang, Fan, Yue, Liang, Yang, and Huang}{Xu
  et~al\mbox{.}}{2018}]%
        {xu2018youtube}
\bibfield{author}{\bibinfo{person}{Ning Xu}, \bibinfo{person}{Linjie Yang},
  \bibinfo{person}{Yuchen Fan}, \bibinfo{person}{Dingcheng Yue},
  \bibinfo{person}{Yuchen Liang}, \bibinfo{person}{Jianchao Yang}, {and}
  \bibinfo{person}{Thomas Huang}.} \bibinfo{year}{2018}\natexlab{}.
\newblock \showarticletitle{Youtube-vos: A large-scale video object
  segmentation benchmark}.
\newblock \bibinfo{journal}{\emph{arXiv preprint arXiv:1809.03327}}
  (\bibinfo{year}{2018}).
\newblock


\bibitem[\protect\citeauthoryear{Xu, Zhang, Yu, Hu, Rowen, Hu, and Shi}{Xu
  et~al\mbox{.}}{2019}]%
        {xu2019dac}
\bibfield{author}{\bibinfo{person}{Xiaowei Xu}, \bibinfo{person}{Xinyi Zhang},
  \bibinfo{person}{Bei Yu}, \bibinfo{person}{X~Sharon Hu},
  \bibinfo{person}{Christopher Rowen}, \bibinfo{person}{Jingtong Hu}, {and}
  \bibinfo{person}{Yiyu Shi}.} \bibinfo{year}{2019}\natexlab{}.
\newblock \showarticletitle{Dac-sdc low power object detection challenge for
  uav applications}.
\newblock \bibinfo{journal}{\emph{IEEE Transactions on Pattern Analysis and
  Machine Intelligence}} (\bibinfo{year}{2019}).
\newblock


\bibitem[\protect\citeauthoryear{Yang, Loquercio, Scaramuzza, and Soatto}{Yang
  et~al\mbox{.}}{2019}]%
        {yang2019unsupervised}
\bibfield{author}{\bibinfo{person}{Yanchao Yang}, \bibinfo{person}{Antonio
  Loquercio}, \bibinfo{person}{Davide Scaramuzza}, {and}
  \bibinfo{person}{Stefano Soatto}.} \bibinfo{year}{2019}\natexlab{}.
\newblock \showarticletitle{Unsupervised moving object detection via contextual
  information separation}. In \bibinfo{booktitle}{\emph{Proceedings of the IEEE
  Conference on Computer Vision and Pattern Recognition}}.
  \bibinfo{pages}{879--888}.
\newblock


\bibitem[\protect\citeauthoryear{Yu, Li, Zhang, Huang, Du, Tian, and Sebe}{Yu
  et~al\mbox{.}}{2019}]%
        {yu2019unmanned}
\bibfield{author}{\bibinfo{person}{Hongyang Yu}, \bibinfo{person}{Guorong Li},
  \bibinfo{person}{Weigang Zhang}, \bibinfo{person}{Qingming Huang},
  \bibinfo{person}{Dawei Du}, \bibinfo{person}{Qi Tian}, {and}
  \bibinfo{person}{Nicu Sebe}.} \bibinfo{year}{2019}\natexlab{}.
\newblock \showarticletitle{The Unmanned Aerial Vehicle Benchmark: Object
  Detection, Tracking and Baseline}.
\newblock \bibinfo{journal}{\emph{International Journal of Computer Vision}}
  (\bibinfo{year}{2019}), \bibinfo{pages}{1--19}.
\newblock


\bibitem[\protect\citeauthoryear{Zhang, Wang, Tian, Gou, and Wang}{Zhang
  et~al\mbox{.}}{2018}]%
        {zhang2018mfr}
\bibfield{author}{\bibinfo{person}{Hui Zhang}, \bibinfo{person}{Kunfeng Wang},
  \bibinfo{person}{Yonglin Tian}, \bibinfo{person}{Chao Gou}, {and}
  \bibinfo{person}{Fei-Yue Wang}.} \bibinfo{year}{2018}\natexlab{}.
\newblock \showarticletitle{MFR-CNN: Incorporating multi-scale features and
  global information for traffic object detection}.
\newblock \bibinfo{journal}{\emph{IEEE Transactions on Vehicular Technology}}
  \bibinfo{volume}{67}, \bibinfo{number}{9} (\bibinfo{year}{2018}),
  \bibinfo{pages}{8019--8030}.
\newblock


\bibitem[\protect\citeauthoryear{Zhu, Du, Wen, Bian, Ling, Hu, Peng, Zheng,
  Wang, Zhang, et~al\mbox{.}}{Zhu et~al\mbox{.}}{2019}]%
        {zhu2019visdrone}
\bibfield{author}{\bibinfo{person}{Pengfei Zhu}, \bibinfo{person}{Dawei Du},
  \bibinfo{person}{Longyin Wen}, \bibinfo{person}{Xiao Bian},
  \bibinfo{person}{Haibin Ling}, \bibinfo{person}{Qinghua Hu},
  \bibinfo{person}{Tao Peng}, \bibinfo{person}{Jiayu Zheng},
  \bibinfo{person}{Xinyao Wang}, \bibinfo{person}{Yue Zhang}, {et~al\mbox{.}}}
  \bibinfo{year}{2019}\natexlab{}.
\newblock \showarticletitle{VisDrone-VID2019: The Vision Meets Drone Object
  Detection in Video Challenge Results}. In
  \bibinfo{booktitle}{\emph{Proceedings of the IEEE International Conference on
  Computer Vision Workshops}}. \bibinfo{pages}{0--0}.
\newblock


\bibitem[\protect\citeauthoryear{Zhu, Wen, Bian, Ling, and Hu}{Zhu
  et~al\mbox{.}}{2018}]%
        {zhu2018vision}
\bibfield{author}{\bibinfo{person}{Pengfei Zhu}, \bibinfo{person}{Longyin Wen},
  \bibinfo{person}{Xiao Bian}, \bibinfo{person}{Haibin Ling}, {and}
  \bibinfo{person}{Qinghua Hu}.} \bibinfo{year}{2018}\natexlab{}.
\newblock \showarticletitle{Vision meets drones: A challenge}.
\newblock \bibinfo{journal}{\emph{arXiv preprint arXiv:1804.07437}}
  (\bibinfo{year}{2018}).
\newblock


\end{thebibliography}

\end{document}